\documentclass[10pt,twocolumn,letterpaper]{article}

\usepackage[pagenumbers]{cvpr} %

\usepackage{graphicx}
\usepackage{amsmath}
\usepackage{amssymb}
\usepackage{booktabs}
\usepackage{multirow}
\usepackage[table]{xcolor} 
\usepackage{xspace}
\usepackage{enumitem}
\usepackage{arydshln}
\usepackage{microtype}
\usepackage[font=small]{caption}
\usepackage{bm}
\usepackage{tabularx} 

\definecolor{newlightblue}{RGB}{0,75,255}
\usepackage[pagebackref=true,breaklinks=true,colorlinks,bookmarks=false,urlcolor={newlightblue},citecolor={newlightblue}]{hyperref}

\usepackage[capitalize]{cleveref}
\crefname{section}{Sec.}{Secs.}
\Crefname{section}{Section}{Sections}
\Crefname{table}{Table}{Tables}
\crefname{table}{Tab.}{Tabs.}

\makeatletter
\def\adl@drawiv#1#2#3{%
        \hskip.5\tabcolsep
        \xleaders#3{#2.5\@tempdimb #1{1}#2.5\@tempdimb}%
                #2\z@ plus1fil minus1fil\relax
        \hskip.5\tabcolsep}
\newcommand{\cdashlinelr}[1]{%
  \noalign{\vskip\aboverulesep
           \global\let\@dashdrawstore\adl@draw
           \global\let\adl@draw\adl@drawiv}
  \cdashline{#1}
  \noalign{\global\let\adl@draw\@dashdrawstore
           \vskip\belowrulesep}}
\makeatother

\DeclareMathOperator*{\argmin}{argmin}

\definecolor{lightyellow}{RGB}{255,255,170}
\definecolor{lightgray}{RGB}{240,240,240}

\newcommand{\ci}[1]{\footnotesize{\textcolor{gray}{~($\pm #1$)}}}

\newcommand{\mypar}[1]{\vspace{-3mm}\paragraph{#1}}
\def\upvspacefig{\vspace{-0.0mm}}
\def\vspacefig{\vspace{-0mm}}
\newcommand{\mysection}[1]{\vspace{0mm}\section{#1}\vspace{0mm}}
\newcommand{\mysubsection}[1]{\vspace{0mm}\subsection{#1} \vspace{0mm}}
\newcommand{\supparxiv}[2]{#2}

\def\cvprPaperID{4478} %
\def\confName{CVPR}
\def\confYear{2023}

\begin{document}

\title{Conditional Generation of Audio from Video via Foley Analogies}

\author{Yuexi Du\textsuperscript{1,2}\qquad Ziyang Chen\textsuperscript{1}\qquad Justin Salamon\textsuperscript{3}\qquad Bryan Russell\textsuperscript{3}\qquad Andrew Owens\textsuperscript{1} \vspace{3.5mm}\\
\textsuperscript{1}University of Michigan\qquad \textsuperscript{2}Yale University\qquad \textsuperscript{3}Adobe Research
}

\twocolumn[{%
\renewcommand\twocolumn[1][]{#1}%
\maketitle
\begin{center}
    \vspace{-3mm}
    
    \includegraphics[width=1.0\textwidth]{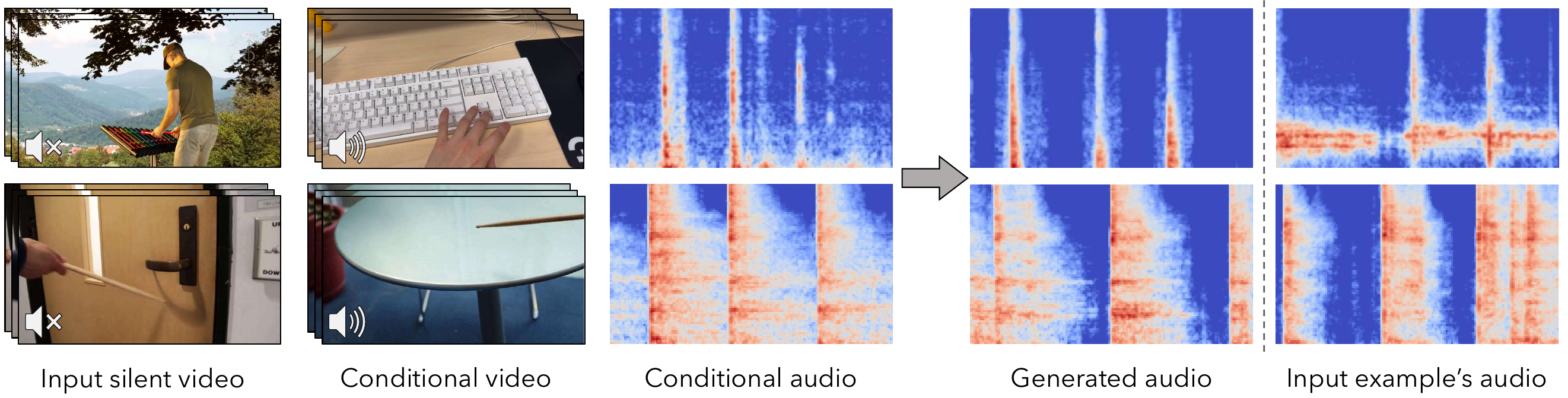}
    \captionof{figure}{{\bf Conditional Foley generation via analogy.} We generate a soundtrack for a silent input video, given a user-provided conditional example specifying what its audio should ``sound like.'' 
    In the first example, we make the xylophone strikes sound like the clicks of a mechanical keyboard.
    In the second, we generate a soundtrack for a video in which the drumstick striking a wooden door sounds as though it were made of metal. Notice that the shape of the sound events in the generated audio~(\eg, thin stripes in the top example) matches the conditional audio and the onsets match the input example's audio. For reference, we provide the input video's (held out) sound on the right. {\bf We encourage the reader to watch and listen to the results on our \href{https://xypb.github.io/CondFoleyGen/}{project webpage}}.}
    \label{fig:teaser}
    \vspace{3mm}
\end{center}%
}]

\begin{abstract}
    The sound effects that designers add to videos are designed to convey a particular artistic effect and, thus, may be quite different from a scene's true sound. Inspired by the challenges of creating a soundtrack for a video that differs from its true sound, but that nonetheless matches the actions occurring on screen, we propose the problem of {\em conditional Foley}. 
    We present the following contributions to address this problem. First, we propose a pretext task for training our model to predict sound for an input video clip using a conditional audio-visual clip sampled from another time within the same source video. 
    Second, we propose a model for generating a soundtrack for a silent input video, given a user-supplied example that specifies what the video should ``sound like''. 
    We show through human studies and automated evaluation metrics that our model successfully generates sound from videos, while varying its output according to the content of a supplied example. Project site: \small{\url{https://xypb.github.io/CondFoleyGen}}.
\end{abstract}

\vspace{-3mm} \mysection{Introduction}

When artists create sound effects for videos, they often ``borrow'' sounds from other sources, then manipulate them to match the on-screen actions.
These artists' aim is not necessarily to convey the scene's true sound, but rather to achieve a desired artistic effect. Thus, the clunk of a coconut shell becomes a trotting horse, or the sizzle of cooking bacon becomes rain\footnote{We encourage you to watch and listen to how sound artists work: \url{https://www.youtube.com/watch?v=UO3N_PRIgX0}}.

The problem of creating sound effects for video, known as Foley~\cite{ament2014foley}, has often been posed as predicting a video's co-occurring sound~\cite{owens2015visually,zhou2018visual,iashin2021taming}. Yet the task that artists solve is subtly different. They create a soundtrack for a video that {\em differs} from its true sound, but that still plausibly matches the on-screen events. Also, these prior systems largely do not give artists control over the output sound.

To aid Foley artists while giving them artistic control, we propose a {\em conditional} Foley problem inspired by classic work on image analogies~\cite{hertzmann2001image}. 
Our task is to generate a soundtrack for an input silent video from a user-provided conditional audio-visual example that specifies what the input video should ``sound like.'' 
The generated soundtrack should relate to the input video in an analogous way as the provided example~(\cref{fig:teaser}). This formulation naturally separates the problem of selecting an exemplar sound, which arguably requires the artist's judgment, from the problem of manipulating that sound to match a video, such as by precisely adjusting its timing and timbre. 

This proposed task is challenging, since a system must learn to adapt the exemplar (conditional) sound to match the timing of the visual content of a silent video while preserving the exemplar sound's timbre. While prior methods can predict a video's sound~\cite{iashin2021taming,owens2015visually,zhou2018visual}, they cannot incorporate an artist's exemplary conditional sound. 
Furthermore, while vision-to-sound methods can pose the problem as predicting a video's soundtrack from its images, it is less clear how supervision for conditional examples can be obtained.

\looseness=-1
To address these challenges, we contribute a self-supervised pretext task for learning conditional Foley, as well as a model for solving it. Our pretext task exploits the fact that natural videos tend to contain repeated events that produce closely related sounds. To train the model, we randomly sample two pairs of audio-visual clips from a video, and use one as the conditional example for the other. 

Our model learns to infer the types of actions within the scene from the conditional example, and to generate analogous sounds to match the input example. At test time, our model generalizes to conditional sounds obtained from other videos. To solve the task, we train a Transformer~\cite{vaswani2017attention} to autoregressively predict a sequence of audio codes for a spectrogram VQGAN~\cite{esser2021taming}, while conditioning on the provided audio-visual example. We improve the model's performance at test time by generating a large number of soundtracks, then using an audio-visual synchronization model~\cite{iashin2022sparse,chung2016out,owens2018audio} to select the sound with the highest degree of temporal alignment with the video.

We evaluate our model on the {\em Greatest Hits} dataset~\cite{owens2015visually}, which contains videos that require an understanding of material properties and physical interactions, and via qualitative examples from the highly diverse {\em CountixAV} dataset~\cite{zhang2021repetitive}. 
Through perceptual studies and quantitative evaluations, we show that our model generates soundtracks that convey the physical properties of conditional examples while reflecting the timing and motions of the on-screen actions.

\mysection{Related Work}

\paragraph{Predicting sound from images and video.}
In early work, Van Den Doel \etal~\cite{van2001foleyautomatic} generated sound for physical simulations. 
More recent examples include predicting soundtracks for videos in which someone strikes objects with a drumstick~\cite{owens2015visually}, generating music from piano~\cite{koepke2020sight}, body motion~\cite{su2020multi}, or dance videos~\cite{gan2020foley,su2021does}, and generating speech from lip motions~\cite{ephrat2017vid2speech,prajwal2020learning}. 
Other work predicted natural sounds (typically ambient sound) using an autoregressive vocoder~\cite{zhou2018visual}, temporal relational networks~\cite{ghose2020autofoley}, and visually guided generative adversarial network~\cite{ghose2021foleygan}. Iashin and Rahtu~\cite{iashin2021taming} recently used a VQGAN~\cite{esser2021taming,van2017neural} operating on mel spectrograms to generate sounds. We adopt this architecture to perform {\em conditioned} sound generation. 
In contrast to previous approaches, our goal is not simply to estimate the sound from a silent video, but to use a user-provided example to tailor the sound to the actions in a scene.

\mypar{Sound design.}
The sounds that occur in a film are often not recorded on-site, but instead are inserted by artists. Sound designers perform a number of steps, including ``spotting" visual events that require sound, choosing or recording an appropriate sound, and manipulating the chosen sound with editing software~\cite{ament2014foley}. Our work addresses this final manipulation step. Other work has sought computational approaches to re-target or match visual signals to audio in sound design problems. Davis and Agrawala~\cite{davis2018visual} re-targeted video to match a soundtrack by aligning both signals at estimated beats. Langlois and James~\cite{langlois2014inverse} addressed the task of synchronizing vision to match sounds. Other work learns to match relevant music with videos~\cite{suris2022time}. However, they do not address the proposed conditional Foley task.

\mypar{Interactive stylization.} 
We take inspiration from the classic work of Hertzmann \etal~\cite{hertzmann2001image}, which learned to restyle input images from a single user-provided example of an image and its stylization.  Like this work, we seek to generalize from one piece of paired data (a video and its sound) to another. 

Zhang \etal~\cite{zhang2017real} learned to colorize images with simulated user-provided hints, using a self-supervised training procedure. In contrast, our model is given user-supplied hints of audio-visual examples from {\em other} videos, rather than from annotations of the input. A variety of methods have been proposed to stylize images with user-provided conditions.  Li \etal~\cite{li2022learning} stylized input images based on user-provided sound. Lee \etal~\cite{lee2021sound} generated images from sound and pretrained language models. Chen \etal~\cite{chen2022visual} learned to alter the acoustics of speech sound to match a visual scene. Many recent methods have applied style transfer to audio. \cite{ulyanov2016audio} These include methods that separate style and content using feature statistics~\cite{ulyanov2016audio,verma2018neural}, following Gatys \etal~\cite{gatys2016image}, and methods that transfer musical timbre~\cite{huang2018timbretron} based on CycleGAN~\cite{zhu2017unpaired}. In contrast, Foley generation requires generating sound without a ground truth audio, since in general there is no existing recorded sound~\cite{ament2014foley} in Foley artists' workflow.  Nistal \etal~\cite{nistal2020drumgan} propose to synthesize drum sounds using a GAN~\cite{goodfellow2014generative} conditioned on audio perceptual features rather than a complete audio-visual conditional clip.

\begin{figure*}[t]
    \centering
    \upvspacefig
    \includegraphics[width=1.0\linewidth]{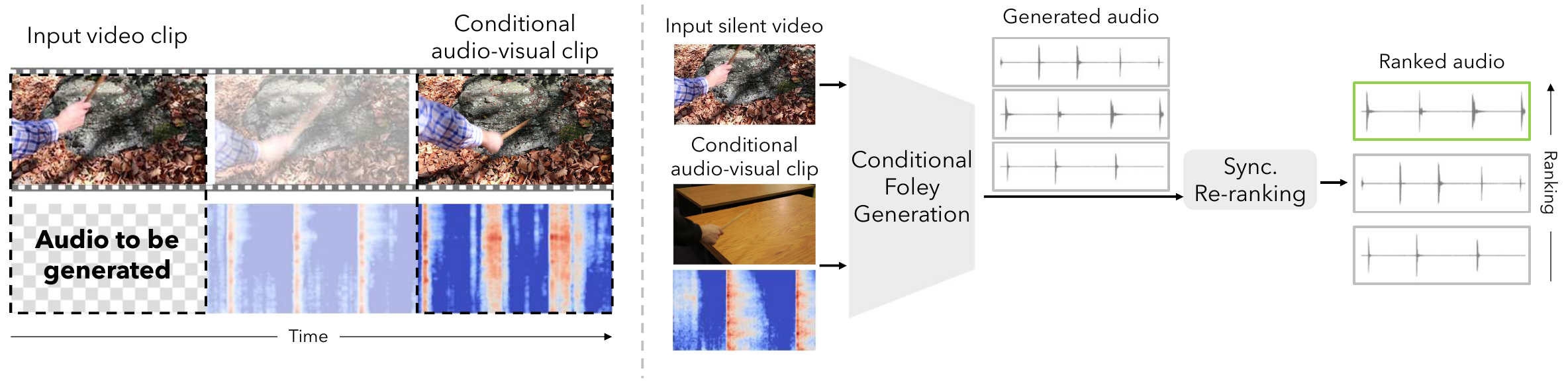}
    \begin{flushleft}
        \vspace{-4.5mm}
        \hspace{14mm} (a) Pretext task for training \hspace{64mm} (b) Inference
         \vspace{-3mm}
    \end{flushleft}
    \caption{{\bf Conditional video-to-audio synthesis via Foley analogy}. (a) For our pretext task, we extract two clips from a longer video and train our model to predict the soundtrack for one, given audio-visual information from the other. Through this process, our model learns to condition its soundtrack predictions on other videos. (b) At test time, we provide the model with a silent input video and an audio-visual clip (taken from another video). We can use an audio-visual synchronization model to re-rank the generated soundtracks and choose the one with the best temporal alignment.}
    \vspacefig
    \label{fig:sampling}
\end{figure*}

\mypar{Self-supervised and few-shot audio-visual learning.}  
Our work aims to learn from data without human annotations. 
There has been much recent progress in learning strong audio-visual representations from video with accompanying audio tracks for other downstream applications~\cite{hershey2001audio,owens2018audio,zhao2018sound,gao2018learning,tzinis2020into,ephrat2018looking,gao20192,garg2021geometry,xu2021visually,yang2020telling,arandjelovic2018objects,chen2021structure,cui2022varietysound}. 
Example applications include using audio-visual data for source separation~\cite{hershey2001audio,owens2018audio,zhao2018sound,gao2018learning,tzinis2020into,ephrat2018looking} and for converting mono sound to stereo~\cite{gao20192,garg2021geometry,xu2021visually,yang2020telling}.
Our work focuses on a different application, but is related to methods that adapt themselves using a small number of labeled examples, such as audio event detection~\cite{wang2020few,wang2021calls}, talking head generation~\cite{chen2020talking}, and diarization~\cite{chung2019said}.

\mysection{Method}

\def\vi{{\mathbf v}_q} %
\def\ai{{\mathbf a}_q} %
\def\vc{{\mathbf v}_c} %
\def\ac{{\mathbf a}_c} %
\def\ag{{\mathbf a}_g} %
\def\bs{{\mathbf s}} %
\def\bz{{\mathbf z}} %
\def\ba{{\mathbf a}} %

Our goal is to generate a soundtrack for a silent input video, given a user-provided {\em conditional} audio-visual example that specifies what the video should ``sound like''. We learn a function $\mathcal{F}_\theta(\vi, \vc, \ac)$ parameterized by $\theta$ that generates a soundtrack from an input video $\vi$, given a conditional video $\vc$ and conditional audio $\ac$. 
We now describe our pretext task for training $\mathcal{F}_\theta$ from unlabeled data, and our conditional vision-to-sound model.

\mysubsection{Pretext task for conditional prediction} 

We desire a pretext task that results in the model obtaining the necessary information from each source. In particular, we would like the input video to specify the type of action (\eg, hitting vs.\ scratching an object) and its timing, while the conditional audio-visual example should specify the timbre of the generated sound (\eg, the type of the materials that are being interacted with).

We define our task as a video-to-audio prediction problem in which another clip from the same video is provided as the conditional example~(\cref{fig:sampling}a). During training, we sample two clips from a longer video, centered at times $t$ and $t + \Delta t$ respectively, using one as the conditional example and the other as the input video. The model is tasked with predicting the sound from the silent input video, using conditional clips as an additional input.

According to this pretext task, we can define a loss $\mathcal{L}$ over an audio target $\ag$ and a prediction $\mathcal{F}_\theta(\vi, \vc, \ac)$ given corresponding input video $\vi$ and conditional audio-visual clip $(\vc, \ac)$:

\begin{equation}
     \mathcal{L}(\ag, \mathcal{F}_\theta(\vi, \vc, \ac))
     \label{eq:loss}
\vspace{-1.5mm}
\end{equation}

This formulation exploits the fact that the actions within a video tend to be closely related~\cite{zhang2021repetitive} (or ``self-similar"~\cite{shechtman2007matching}), such as when an action is performed repeatedly. Thus randomly sampled pairs of clips frequently contain related actions. When this occurs, the model can use conditional sound to improve its prediction. However, the model cannot solve the task by simply ``copying and pasting" the conditional sound, since it must account for the content of the timing of actions (and the type of motion) in the input video. 
Since the model is trained to assume that the conditional example is informative about the input, we empirically find that it learns to base its prediction on the conditional sound.  
This finding allows for substituting in a conditional sound sampled from a completely {\em different} video at test time (\cref{fig:sampling}b).

\begin{figure*}[t]
    \centering
    \upvspacefig
    \includegraphics[width=1.0\linewidth]{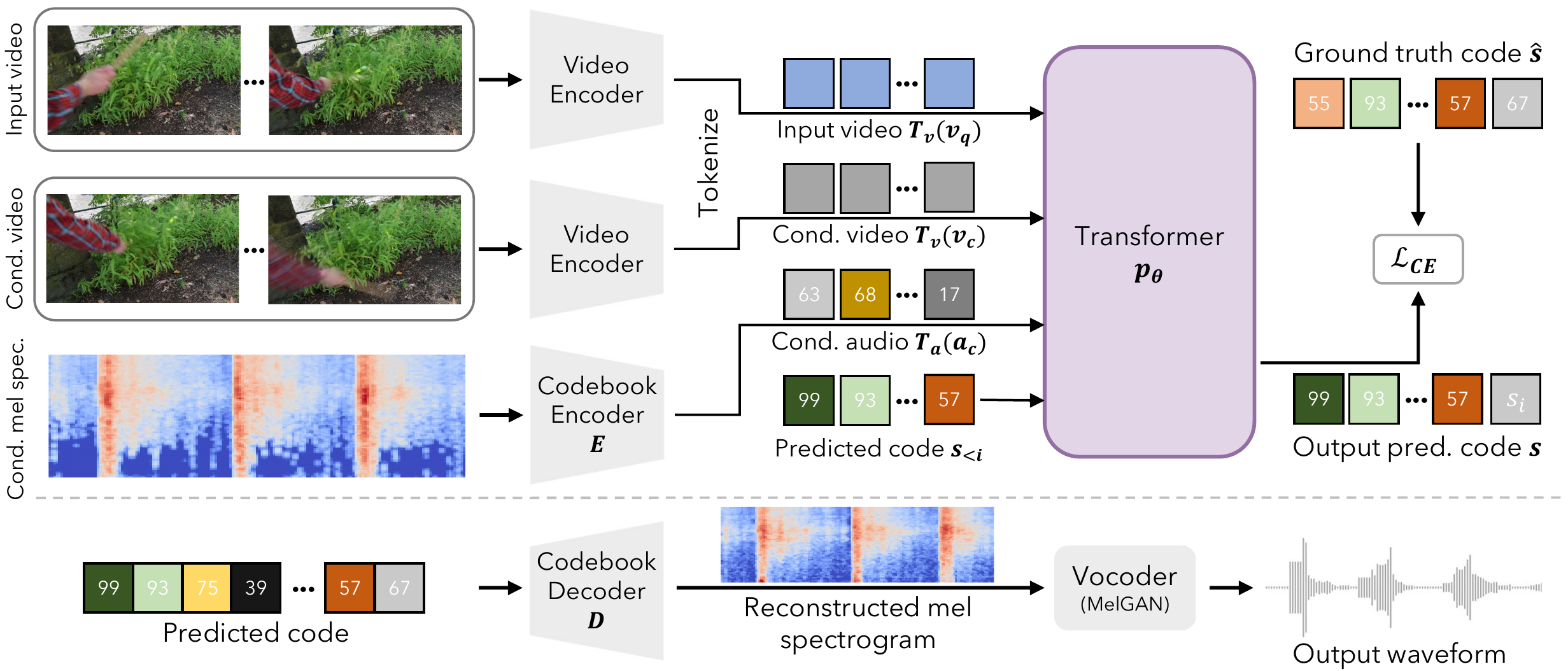}
    \caption{{\bf Conditional Foley generation}. (Top) We predict the soundtrack for a silent video, conditioned on an audio-visual pair sampled from the same video. We encode and tokenize the video and audio signals, and feed them into a transformer. This transformer autoregressively predicts a code from a VQGAN~\cite{esser2021taming,iashin2021taming}, representing the input example's sound. (Bottom) We generate a waveform by converting the code to a mel spectrogram, then using a MelGAN~\cite{kumar2019melgan} vocoder to convert it to a waveform. Here, $\mathcal{L}_{CE}$ represents a cross-entropy loss.}
    \vspacefig
    \label{fig:method}
\end{figure*}

\mysubsection{Conditional sound prediction architecture} 
\label{soundpred}

We describe our architecture of $\mathcal{F}_\theta$ for, first, obtaining a code representation for a target sound via VQGAN~\cite{esser2021taming,iashin2021taming}  and, second, predicting an output sound for a given input video and conditional audio-visual pair.

\mypar{Vector-quantized audio representation.} We follow Iashin and Rahtu ~\cite{iashin2021taming} and represent the predicted sound as a sequence of discrete codes, using a VQGAN~\cite{esser2021taming,van2017neural} that operates on mel spectrograms\footnote{We use {\em log} mel spectrograms unless otherwise noted.}. We learn this code by training an autoencoder to reconstruct sounds in a dataset, using the codes as its latent vector.  After training, a predicted code sequence can subsequently be converted to a waveform.

Given a waveform $\ba$ and its mel spectrogram $\mathrm{MSTFT}(\ba) \in \mathbb{R}^{T \times F}$, we compute embeddings ${\hat \bz} = E(\mathrm{MSTFT}(\ba)) \in \mathbb{R}^{T' \times F' \times d}$, where $T'$ and $F'$ define a lower-resolution time-frequency grid, $d$ is the dimensionality of the embedding at each patch, and $E$ is a CNN. Each embedding vector is then replaced with the nearest entry in a codebook $\{\mathbf{c}_k\}_{k=1}^K$:

\begin{equation}
     {\bz}_{tf} = q(\hat \bz_{tf}) = \argmin_{\mathbf{c}_k} ||\hat{\bz}_{tf} - {\bf c}_k||,
     \label{eq:vq}
\vspace{-1.5mm}
\end{equation}
where $\hat{\bz}_{tf}$ is the embedding at time-frequency index $(t, f)$. We train the model to reconstruct the input sound $\hat {\mathbf S} = D(q(E(\mathrm{MSTFT}(\ba))))$, where $D$ is a CNN-based decoder and $q$ is applied to every embedding. We use the loss function from~\cite{iashin2021taming}, which adapts the VQGAN loss~\cite{esser2021taming} to spectrograms, jointly minimizing a mean-squared error reconstruction loss~\cite{van2017neural},  a perceptual loss~\cite{johnson2016perceptual}, and a patch-based discriminator loss~\cite{isola2017image}. We provide details in \supparxiv{the supp.}{\cref{appendix:loss}.}

Finally, we can obtain a {\em code} $s \in \{0, 1, \cdots, K-1\}^{T' \times F'}$ for a sound from the VQGAN by replacing each quantized vector in $\bz$ with the index of its nearest codebook entry, \ie, $s_{tf}$ is the index of the selected codebook entry in \cref{eq:vq}.

\mypar{Autoregressive sound prediction.}
With the predicted code $s$, we can now formulate the likelihood of generating code $s$ from the silent input video and the conditional example. We order the indices of $s$ in raster scan order~\cite{esser2021taming} and predict them autoregressively: 
\begin{equation}
    p_\theta(s | \vi, \vc, \ac) = \prod_i p_\theta(s_i | s_{< i}, \vi, \vc, \ac),
    \label{eq:autoregression}
\end{equation}
where $s_{< i}$ are the previous indices in the sequence. Given these probabilities, we formulate $\mathcal{L}$~(\cref{eq:loss}) as the cross-entropy loss between the predicted token $s_i$ and ground-truth token $\hat{s_i}$.

Having defined a code-based representation for sounds, we describe our architecture of $\mathcal{F}_\theta$ for conditional sound prediction~(\cref{fig:method}).  Following~\cite{esser2021taming,iashin2021taming}, we predict the code sequence~(\cref{eq:autoregression}) using a decoder-only transformer~\cite{vaswani2017} based on GPT-2~\cite{radford2019language}. The inputs to this transformer are tokenized versions of $\vi, \vc$, and $\ac$. We now describe how these signals are converted into tokens. 

\mypar{Input representations.} We represent each video signal using a ResNet (2+1)D-18~\cite{tran2018}. To preserve fine-grained temporal information, we remove all temporal striding, so that the final convolutional layer has the same temporal sampling rate as the input video. We perform average pooling over the spatial dimension, resulting in an embedding vector for each frame. Each such vector becomes a token. We denote this tokenization operation $T_v({\mathbf v})$.

We represent the conditional audio signal using its vector-quantized embeddings. Specifically, we compute $\bz^{(c)} = q(E(\mathrm{MSTFT}(\ac))) \in \mathbb{R}^{T' \times F' \times d}$~(\cref{eq:vq}) and extract its $d$-dimensional embedding vectors $\bz_1^{(c)}, \bz_2^{(c)}, \cdots, \bz_N^{(c)}$ in raster-scan order. We denote this tokenization operation $T_a(\ac)$. 

We combine these tokens into a single sequence: $\mathcal{S} = \mathrm{Concat}(T_v(\vc), T_v(\vi), T_a(\ac))$. Thus, we model $p_\theta(s | \vi, \vc, \ac) = p_\theta(s | \mathcal{S})$. Following standard practice~\cite{esser2021taming,van2017neural}, we generate the audio code autoregressively, feeding the previously generated codes back into the model using their vector-quantized representation.

\begin{table*}[t!]
\upvspacefig
\centering
\setlength{\tabcolsep}{8pt}
\resizebox{\textwidth}{!}{
\begin{tabular}{lcccccccccc}
\toprule
\multirow{4}{*}{Model} & \multicolumn{10}{c}{Task}   \\ 
\cmidrule{2-11} 
& \multicolumn{3}{c}{Material}            &                  & \multicolumn{3}{c}{Action}  &   & \multicolumn{2}{c}{Onset}                               \\
    & match             & mismatch          & overall   &         & match             & mismatch          & overall   &     & \# onset   &    onset sync.      \\
& Acc (\%)          & Acc (\%)          & Acc (\%)   &        & Acc (\%)          & Acc (\%)          & Acc (\%) &   & Acc (\%) & AP (\%)       \\ 
\midrule
Style transfer\textsuperscript{$\ast$}~\cite{ulyanov2016audio,gatys2015neural} &  30.0             &  33.5             &  32.3       &       &  20.8             &  36.6             &  31.3   &   &    19.1  &     46.9     \\
Onset transfer                                          & \textbf{54.8} & \textbf{51.4} & \textbf{52.6}  &  & \textbf{69.0} & \textbf{44.7}   & \textbf{52.9}   & & \textbf{24.8}  & \textbf{71.9}  \\ 
\midrule
Chance                                  &  5.9  &  5.9  &  5.9  &  &  50.0 &  50.0  &  50.0   &  & -- & --\\
SpecVQGAN~\cite{iashin2021taming}                       &  25.4             &  26.8             &  26.1         &    &  52.3             &  43.1             &  46.2   &  &  11.3      &  51.0      \\
SpecVQGAN - finetuned~\cite{iashin2021taming}                       &  29.9             &  25.7             &  27.2         &    &  70.6             &  58.4             &  62.5   &  &  25.8      &  59.3      \\
Ours - No cond.                                         &  21.3             &  24.9             &  23.7     &        &  61.4             &  55.1             &  57.2     &    &  24.6   &  59.3     \\
Ours - Base      &   41.1   &   41.6   &  41.4  &  &  67.5   &  59.2  &  62.0  &  & \textbf{26.5}  & \textbf{60.0}  \\ 
Ours - w/ re-rank                                         & \textbf{43.4}             & \textbf{45.2}            & \textbf{44.0}     &        & \textbf{78.2}             & \textbf{61.3}             & \textbf{66.7}     &    &  25.3   &  54.3     \\

\bottomrule
\end{tabular}
}
\caption{{\bf Automated evaluation metrics.} We measure the rate at that generated sounds have the material properties of the conditional examples, the actions of the input examples, and the number and the timing of the onsets in the generated sound with respect to the original sound. We further break down the automated metrics according to whether the conditional and input examples have the {\bf matched} (or {\bf mismatched}) actions and materials. The number of onsets is measured by whether the generated sound has the same number of onsets as the original sound. We measure the average precision of onset predictions that are within 0.1 seconds of the ground truth to evaluate the timing of the generated onsets. $\ast$ indicates that the model is an ``oracle'' and accesses the input example's sound. }

\vspacefig
\label{tab:VGGishTestMerged}
\end{table*}

\mypar{Generating a waveform.} 
Our complete model $\mathcal{F}_\theta$ works by first generating a code using a transformer, converting it to a mel spectrogram using the decoder $D$, then converting the mel spectrogram to a waveform. To perform this final step, we follow~\cite{iashin2021taming} and use a pretrained MelGAN vocoder~\cite{kumar2019melgan}. We found that this produced significantly better results than standard Griffin-Lim~\cite{griffin1984signal}.

\mypar{Re-ranking based on audio-visual synchronization.} Inspired by other work in cross-modal generation, we use {\em re-ranking} to improve our model's predictions~\cite{ramesh2021zero}. We generate a large number of sounds, then select the best one, as judged by a separate classifier. Typically, these approaches use a model that judges the multimodal agreement between the input and output. In our case, however, such a classifier ought to consider conditionally  generated sound to be a poor match for both the input and conditional videos. We instead propose to use an {audio-visual synchronization} model~\cite{chung2016out,owens2018audio,iashin2022sparse} to measure the temporal alignment between the predicted sound and the input video. These models predict a temporal offset that best aligns visual and audio data.

As shown in \cref{fig:sampling}b, we use an off-the-shelf synchronization model~\cite{iashin2022sparse} to estimate the offset $t$ between the audio and video and the prediction's confidence. We find the minimum absolute offset $\min |t|$ among all outputs. Then the outputs with an absolute offset greater than $\min |t| + \tau$ are removed, where $\tau$ is the offset tolerance. Finally, we select the sound with the highest confidence.

\begin{figure*}[t]
    \centering
    \upvspacefig
    \includegraphics[width=1.0\linewidth]{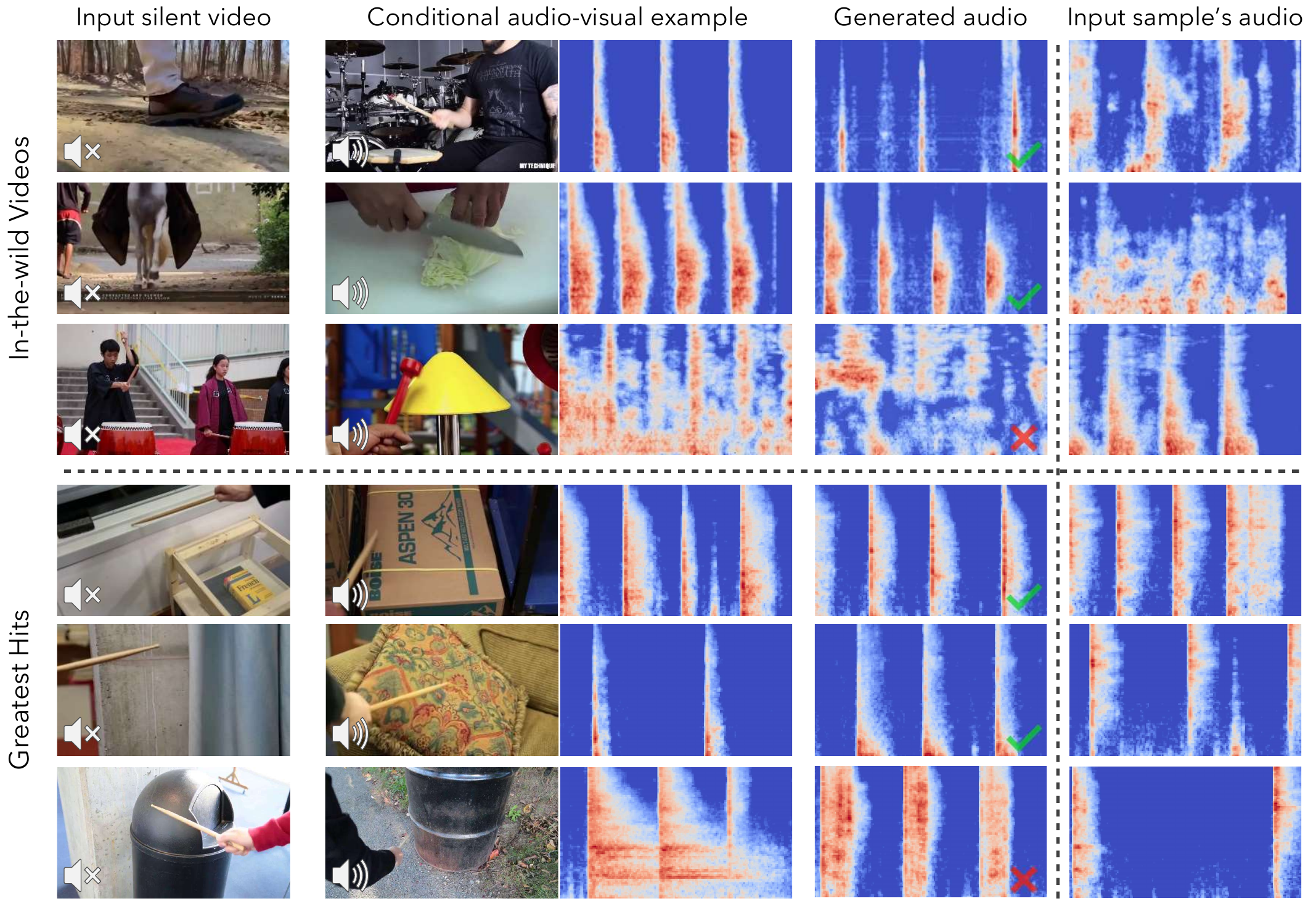}
    \vspace{-4mm}
    \caption{{\bf Qualitative results}. We present results from our model. We show the result for the internet videos from the model trained on {\em CountixAV} dataset~\cite{zhang2021repetitive}~(Row 1--3) and the {\em Greatest Hits} dataset~\cite{owens2015visually}~(Row 4--6). Rows 3 and 6 show failure cases with red crosses on generated audio. The timing and number of hits in the generated audio largely match that of the (held out) audio for the input video, suggesting the generated audio is matching the actions in the input video. The frequencies of the generated audio approximately match the conditional example, indicating a similar timbre. {\bf To hear the sounds and see more examples, please refer to our \href{https://xypb.github.io/CondFoleyGen/}{project webpage}}.}   
    \vspacefig
    \label{fig:result}
\end{figure*}

\mysection{Experiments}

To evaluate our method, we use a combination of automatic evaluation metrics and human perceptual studies. 

\mysubsection{Experiment Setup}
\vspace{3mm}

\mypar{Dataset.} We train our conditional Foley generation model on datasets of video clips: {\em Greatest Hits}~\cite{owens2015visually}, which is composed of videos of a drumstick interacting with different objects in scenes, and {\em CountixAV}~\cite{zhang2021repetitive}, which contains videos with as much as 23 different classes of repeated actions from in-the-wild YouTube video. These are challenging datasets for audio generation since they require precise timing, and varying sounds subtly based on fine-grained visual properties. In particular, the {\em Greatest Hits} task requires an understanding of the motion of the drumstick and the material properties of the objects. Since this dataset is straightforward to analyze in terms of actions and materials, we use it for our quantitative evaluation, while for {\em CountixAV} we provide qualitative results with permission to evaluate with similar videos in the wild. We provide more information about the dataset and the implementation details in \supparxiv{the supplement}{\cref{appendix:dataset} and \cref{appendix:implementation}}.

\mypar{Other models.} 
We consider a variety of other models for comparison. First, we use the {\bf SpecVQGAN} model of Iashin and Rahtu~\cite{iashin2021taming}, a state-of-the-art vision-to-sound prediction method based on a two-stream visual network. We use the publicly released implementation and weight. We also finetuned the model on the {\em Greatest Hits}~\cite{owens2015visually} dataset for the automated metrics as a fair comparison.

We also consider several ablations of our model:
{\bf No conditional example:} We remove all conditional information from the model. This model is a vision-to-sound prediction method that resembles SpecVQGAN~\cite{iashin2021taming} after controlling for architectural and data variations from our model. {\bf No conditional video:} This model is provided with the conditional audio $\ac$ but not the video $\vc$, and hence cannot observe how the audio and visual events are connected in the conditional video. {\bf No augmentation: } A model trained without audio augmentation. {\bf Random conditional examples: }  A model that is trained with conditional audio-visual clips that are unrelated to the input video, and hence uninformative. We select these clips randomly from other videos in the dataset. {\bf Re-ranked examples: } We generate 100 outputs for each pair of input and condition, then re-rank them.

To better understand our model's behavior, we compare it against two ``non-generative'' methods. First, we propose a model called {\bf Onset Transfer}. Instead of generating the sound, as our model does, this model uses a hand-crafted approach for transferring sounds from the conditional example. We train a ResNet (2+1)-D~\cite{tran2018} model to detect audio onsets from video in both the conditional and input videos, then transfer sounds extracted from random onsets in the conditional example (see \supparxiv{the supp.}{\cref{appendix:onset}} for more details). Second, we evaluated an audio {\bf Style Transfer} method. We used the model of Ulyanov~\cite{ulyanov2016audio}, which applies the stylization method of Gatys \etal~\cite{gatys2015neural} to spectrograms. We note that this method {\em requires audio} as input and is not designed for Foley generation. To address this, we provide the model with $\ai$, the ground truth audio from the input video, thus giving it oracle information.

\mysubsection{Automated Timbre Evaluation}

A successful prediction method should accurately convey the actions in the input video but the material properties of the conditional example. To evaluate whether this is the case, trained classifiers to recognize the action (hit vs. scratch) and the material (a 17-way classification problem), using the labels in the {\em Greatest Hits} dataset~\cite{owens2015visually}. We then used it to classify the predicted, conditional, and input sounds, and compared the estimated labels.

\mypar{Sound classifier.} We finetune a pretrained VGGish classifier~\cite{hershey2016cnn, hershey2017cnn} on the {\em Greatest Hits}~\cite{owens2015visually} dataset to recognize the action or the material from a mel spectrogram. To avoid ambiguity, we only used clips that contained a single material or action type. We provide more details in \supparxiv{the supplement}{\cref{appendix:soundclassifier}}.

\mypar{Evaluation metrics.} We used two evaluation metrics that capture our two criteria: {\em action accuracy}, the fraction of predicted sounds that have the same estimated action category as the~(held out) input sound, and {\em material accuracy}, the fraction of predicted sounds that have the same estimated material category as the conditional example.

\mypar{Results.} We found (\cref{tab:VGGishTestMerged}) that our model performed significantly better on the {\em material} metric than SpecVQGAN and than the variation without conditional examples, both of which are unconditional vision-to-sound prediction methods. As expected, the onset transfer model obtains near-optimal performance, since it simply transfers sounds from the conditional sound, which (trivially) are likely to have the same estimated category. On the other hand, this onset transfer baseline performs poorly on the {\em action} metric, since it has no mechanism for adapting the transferred sounds to the actions in the video (\eg, converting hits to scratches). By contrast, our model obtains high performance on this metric.

To further understand the source of performance differences, we broke down the results according to whether the properties of the conditional example matched those of the input sound or not (\cref{tab:VGGishTestMerged}). As expected, the onset transfer method performs strongly on material metrics, since ``copy and pasting'' conditioning sounds is a trivial solution. On the other hand, our generative approaches significantly outperform it when there is a mismatch between action types, since they can adapt the sound to match the action.

Additionally, we found that the synchronization re-ranking significantly boosted performance on both material and action tasks~(\cref{tab:VGGishTestMerged}). The re-ranked model outperforms the onset transfer method on all three action-related metrics. It also narrows the gap to the onset transfer method for material-related metrics. This demonstrates the effectiveness of the re-ranking method, as well as the advantage of posing our approach as a generative model.

\begin{figure*}[t]
    \centering
    \upvspacefig
    \includegraphics[width=1.0\linewidth]{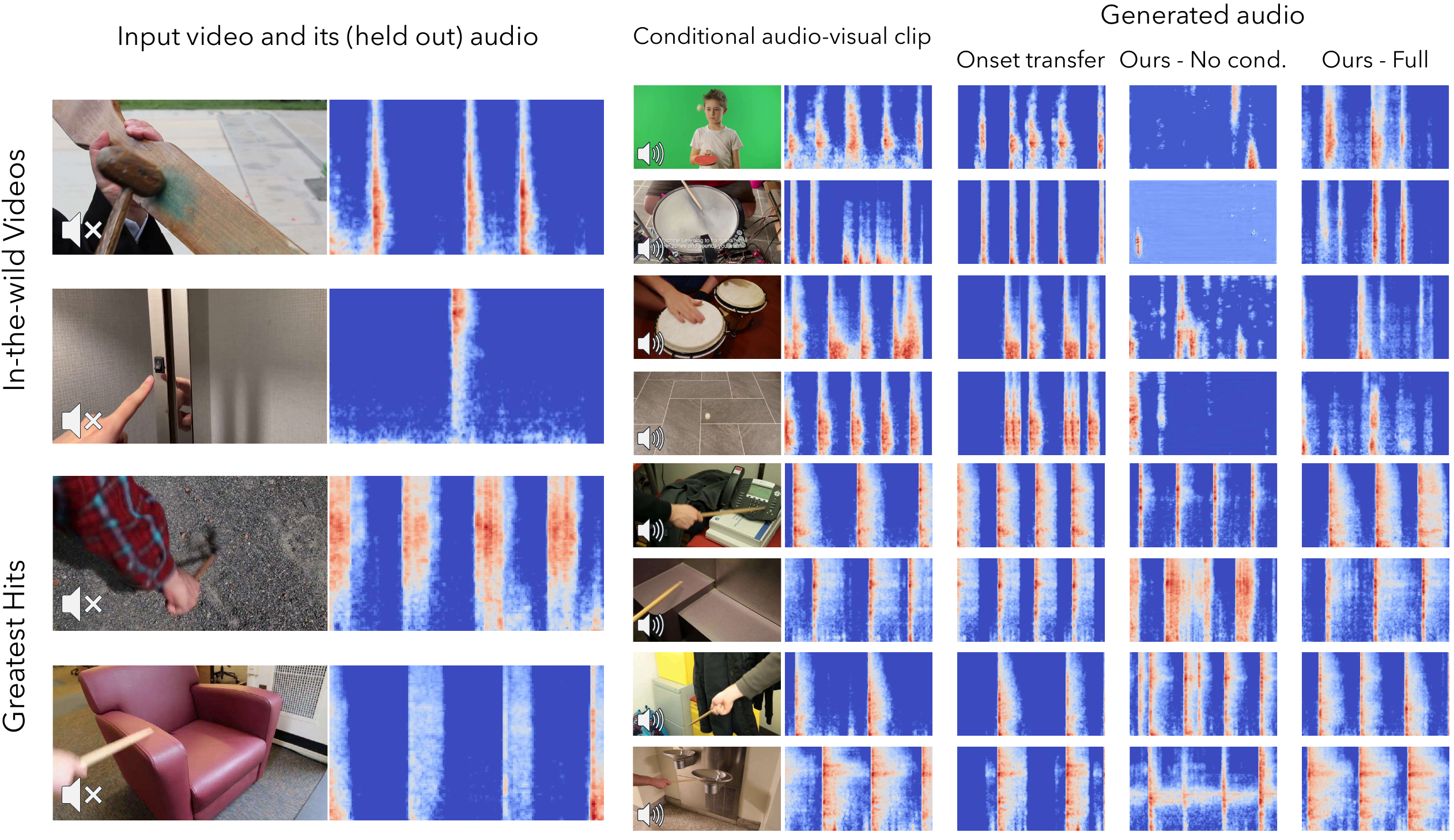} 
    \caption{{\bf Model comparison}. We show conditional Foley generation results for several models, using multiple conditional sounds. We show the result for publicly-sourced demonstration videos (Row 1--2) and the {\em Greatest Hits} dataset~\cite{owens2015visually}~(Row 3--4). Each of the input videos is paired with two different conditional videos. We provide 2 samples from the model variation with no conditional example.}
    \vspacefig
    \label{fig:qualitatve}
\end{figure*}

\mysubsection{Automated Onset Evaluation} %

To evaluate the quality of the generated timbre, we seek to evaluate whether the conditional Foley generation model generates sounds whose onsets match those of the (held-out) sound in the input video. We evaluate two criteria: whether the generated sound contains the correct number of onsets, and whether their timing matches those in the input video.

\mypar{Evaluation metrics.} We measure the fraction of video clips that contain the same number of onsets as the ground truth audio. Following Owens \etal~\cite{owens2016visually}, we report the average precision of detecting the correct onset where the relative wave amplitude provides the confidence of each onset. A detection is correct if it lies within a 0.1-second window of the ground truth.

\mypar{Results.} We notice (\cref{tab:VGGishTestMerged}) that our model outperforms all the generative baselines and the Style transfer method in both metrics. It is not surprising that the onset transfer method obtains the best performance on the onset synchronization task, given that it is explicitly trained on a closely related task. The success of our model in generating the correct number of onset compared with the onset transfer model can be explained by the extra information about the audio-visual relation provided in the condition, which helps the model better understand the action in the video. As shown in \cref{tab:VGGishTestMerged}, the accuracy in capturing the correct number of onsets drops to the same level as the onset transfer method if we remove conditional information. Interestingly, we find (\cref{tab:VGGishTestMerged}) a drop in the performance for the re-ranked model. This may be due to the domain shift from the {\em VGG-Sound} dataset~\cite{chen2020vggsound} that the synchronization network~\cite{iashin2022sparse} was trained on, which may make it difficult to infer the precise timing of the sounds (rather than an overall assessment of whether the two streams are synchronized).

\begin{table}[t]
\centering
\setlength{\tabcolsep}{3pt}
\resizebox{\columnwidth}{!}{
\begin{tabular}{llcc}
\toprule
\multirow{3}{*}{Model}   & \multirow{3}{*}{Variation}  & \multicolumn{2}{c}{Task}                                  \\
\cmidrule{3-4} 
            &                                 & Material                      & Sync.           \\
            &            & Chosen(\%)~$\uparrow$                 & Chosen(\%)~$\uparrow$                 \\ \midrule
Style transfer\textsuperscript{$\ast$}~\cite{ulyanov2016audio,gatys2015neural} & --  &  9.9 \ci{2.3}          &  10.6 \ci{2.4}          \\ %
Onset transfer                                                              & -- & \textbf{64.7 \ci{3.8}} & \textbf{57.3 \ci{3.9}} \\
\midrule
SpecVQGAN~\cite{iashin2021taming}       &      --        &  16.3 \ci{2.9}           &  18.0 \ci{3.0}          \\
\cdashlinelr{1-4}
\multirow{6}{*}{Ours}       &   base     &  50.0 \ci{0.0}          &  50.0 \ci{0.0}   \\
                            &  -  cond. &  35.3 \ci{3.8}          &  40.1 \ci{4.0}     \\
                          &  -  cond. video        &  46.1 \ci{4.0}          &  45.0 \ci{4.0}          \\
                          &  -  augment   &  51.3 \ci{3.9}          &  49.7 \ci{4.0}         \\
                          & w/ rand. cond.  &  45.5  \ci{4.0}          &  47.5 \ci{4.0}          \\
                          &  +  re-rank   & \textbf{54.3 \ci{3.4}}          & \textbf{53.8 \ci{3.4}}          \\ 
\bottomrule
\end{tabular}
}
\caption{{\bf Perceptual study results.} We report the rate at which participants chose a given method's results over our method~(base) for the two questions in our study. For reference, we include the rate that our base method would obtain in the study~(50\%). We report results in terms of 95\% confidence interval. }
\label{tab:HumanTest}
\vspacefig
\end{table}
\mysubsection{Perceptual Study}

We also evaluated our model using a perceptual study, conducted using Amazon Mechanic Turk. We provide the participant with the conditional audio-visual clip and two input videos whose sound was generated by different models (our base model, plus a randomly chosen alternative). 

We asked the participants to judge the generated sounds on two criteria that are similar to the automated metrics. Participants were asked to select: 1) in which result the audio is better synchronized with the actions in the video, 2) in which result the sound is most like that of the object or material in the conditional example. The 376 participants in our study were shown 21 sets of videos, randomly sampled from the evaluation set, the first 5 of which were used as practice and not counted.

\mypar{Comparison to other variations of our model.} We evaluate the influence of different variations of our model in terms of choosing the corresponding method in the perceptual study (\cref{tab:HumanTest}). Our re-ranked model performs best overall and the base model without re-ranking also beats most of the baselines on both metrics. The model with no conditional example obtains poor performance on the material metric but obtains a relatively smaller decrease in synchronization performance. This is understandable, since the model did not use the conditional example but still was encouraged to be synchronized with the video. The model with no conditional video incurs a small drop in the material metrics and a relatively larger drop in synchronization, perhaps because it is unable to observe the relationship between images and sound in the conditional example.

\mypar{Comparison to other approaches.} We compare our model to other methods. Overall, the variation with no conditional example (which is a vision-to-sound method) outperforms SpecVQGAN~\cite{iashin2021taming}. The style transfer~\cite{ulyanov2016audio,gatys2015neural} model performs poorly (and qualitatively often contains artifacts). Interestingly, the onset transfer model performs quite well on the perceptual study, outperforming our base model and the model with re-ranking, despite the fact that it does not tailor its output to the actions in the scene (\cref{tab:VGGishTestMerged}). This is understandable, since it ``copy and pastes" sounds from the conditional example at times that are chosen to be synchronized with impacts. Thus, the user is likely to observe that it exactly matches the conditional sounds and (trivially) conveys the same properties. However, this only occurs when the audio events are cleanly separated in time, and we expect the model to fail when sounds are not easily divided into discrete onsets, or when onsets are ambiguous.

\mysubsection{Qualitative Results} We provide visualizations of predicted sounds from our test set in \cref{fig:result}. Through this visualization, we can see that our model successfully generates sounds that resemble spectral properties of the conditional sound, while matching the timing and actions in the input video's (held out) sound. 
We also present the qualitative result (\cref{fig:result}) of our method on the wild videos from the model trained on the {\em CountixAV} dataset~\cite{zhang2021repetitive}. We follow the same training and generation scheme as for the {\em Greatest Hits}~\cite{owens2015visually}. The model again generates sounds with matching conditional spectral features and input action synchronization.

In \cref{fig:qualitatve}, we visualize how our results vary as a function of the conditional sound and compare our model with the baselines on both datasets. For each input video, we show predictions from different models. We see that our model varies its output depending on the conditional sound (\eg, varying its output based on whether the conditional example is plastic or metal). We see that the onset transfer method ``copies and pastes'' sounds from the conditional example at the correct times. We also observe the failure of the onset transfer baseline (see \cref{fig:qualitatve} Video in the wild part) in a more realistic scenario, where actions and sound in it are more abundant and complex. We show two random samples from the model with no conditioning. We see that the prediction generally matches the input sound, rather than the conditioning sound, and that there are large amounts of variation in the generated audio's timbre. Please refer to \supparxiv{the supp.}{our \href{https://xypb.github.io/CondFoleyGen/}{project webpage}} to listen to our outputs and for more qualitative results.

\mysection{Discussion}

In this paper, we proposed the {\em conditional Foley} task. We also proposed a method for solving this problem through self-supervised learning. We evaluated our method on the {\em Greatest Hits} dataset, finding through perceptual studies and automated metrics that our model successfully learns to transfer relevant information from a conditional sound, while matching the events within the silent input video. We also demonstrate the effectiveness of the model on more complex and realistic data from publicly-sourced videos with the model trained on the {\em CountixAV} dataset.

We see our work potentially opening several directions. Our work tackled one step of the sound design process---the process of manipulating sound to match a video. We see this as a step toward the broader goal of semi-automated  ``user in the loop'' sound design. We also see our work as a step toward synthesis methods that can learn by analogy, in the tradition of classic work such as image analogies~\cite{hertzmann2001image}. We will release code and models on our \href{https://xypb.github.io/CondFoleyGen/}{project site}.

\mypar{Limitations and Broader Impacts.} While soundtrack generation is useful for creative applications, such as film making, it can also be used to create videos that can potentially be used to create disinformation, which can have negative outcomes. The field of image and audio forensics can help mitigate this outcome.

\mypar{Acknowledgements.}  
We thank Jon Gillick, Daniel Geng, and Chao Feng for the helpful discussions. This work was funded in part by DARPA Semafor and Cisco Systems, and by a gift from Adobe. The views, opinions and/or findings expressed are those of the authors and should not be interpreted as representing the official views or policies of the Department of Defense or the U.S. Government.

{\small
\bibliographystyle{ieee_fullname}
\bibliography{avanalogy}

\begin{thebibliography}{10}\itemsep=-1pt

\bibitem{ament2014foley}
Vanessa~Theme Ament.
\newblock {\em The Foley grail: The art of performing sound for film, games,
  and animation}.
\newblock Routledge, 2014.

\bibitem{arandjelovic2018objects}
Relja Arandjelovic and Andrew Zisserman.
\newblock Objects that sound.
\newblock In {\em Proceedings of the European conference on computer vision
  (ECCV)}, pages 435--451, 2018.

\bibitem{chen2022visual}
Changan Chen, Ruohan Gao, Paul Calamia, and Kristen Grauman.
\newblock Visual acoustic matching.
\newblock {\em arXiv preprint arXiv:2202.06875}, 2022.

\bibitem{chen2020vggsound}
Honglie Chen, Weidi Xie, Andrea Vedaldi, and Andrew Zisserman.
\newblock Vggsound: A large-scale audio-visual dataset.
\newblock In {\em ICASSP 2020-2020 IEEE International Conference on Acoustics,
  Speech and Signal Processing (ICASSP)}, pages 721--725. IEEE, 2020.

\bibitem{chen2020talking}
Lele Chen, Guofeng Cui, Celong Liu, Zhong Li, Ziyi Kou, Yi Xu, and Chenliang
  Xu.
\newblock Talking-head generation with rhythmic head motion.
\newblock In {\em European Conference on Computer Vision}, pages 35--51.
  Springer, 2020.

\bibitem{chen2021structure}
Ziyang Chen, Xixi Hu, and Andrew Owens.
\newblock Structure from silence: Learning scene structure from ambient sound.
\newblock In {\em 5th Annual Conference on Robot Learning}, 2021.

\bibitem{chung2019said}
Joon~Son Chung, Bong-Jin Lee, and Icksang Han.
\newblock Who said that?: Audio-visual speaker diarisation of real-world
  meetings.
\newblock {\em arXiv preprint arXiv:1906.10042}, 2019.

\bibitem{chung2016out}
Joon~Son Chung and Andrew Zisserman.
\newblock Out of time: automated lip sync in the wild.
\newblock In {\em Asian conference on computer vision}, pages 251--263.
  Springer, 2016.

\bibitem{cui2022varietysound}
Chenye Cui, Yi Ren, Jinglin Liu, Rongjie Huang, and Zhou Zhao.
\newblock Varietysound: Timbre-controllable video to sound generation via
  unsupervised information disentanglement.
\newblock {\em arXiv preprint arXiv:2211.10666}, 2022.

\bibitem{davis2018visual}
Abe Davis and Maneesh Agrawala.
\newblock Visual rhythm and beat.
\newblock In {\em Proceedings of the IEEE Conference on Computer Vision and
  Pattern Recognition Workshops}, pages 2532--2535, 2018.

\bibitem{ephrat2018looking}
Ariel Ephrat, Inbar Mosseri, Oran Lang, Tali Dekel, Kevin Wilson, Avinatan
  Hassidim, William~T Freeman, and Michael Rubinstein.
\newblock Looking to listen at the cocktail party: A speaker-independent
  audio-visual model for speech separation.
\newblock {\em SIGGRAPH}, 2018.

\bibitem{ephrat2017vid2speech}
Ariel Ephrat and Shmuel Peleg.
\newblock Vid2speech: speech reconstruction from silent video.
\newblock In {\em 2017 IEEE International Conference on Acoustics, Speech and
  Signal Processing (ICASSP)}, pages 5095--5099. IEEE, 2017.

\bibitem{esser2021taming}
Patrick Esser, Robin Rombach, and Bjorn Ommer.
\newblock Taming transformers for high-resolution image synthesis.
\newblock In {\em Proceedings of the IEEE/CVF Conference on Computer Vision and
  Pattern Recognition}, pages 12873--12883, 2021.

\bibitem{gan2020foley}
Chuang Gan, Deng Huang, Peihao Chen, Joshua~B Tenenbaum, and Antonio Torralba.
\newblock Foley music: Learning to generate music from videos.
\newblock In {\em European Conference on Computer Vision}, pages 758--775.
  Springer, 2020.

\bibitem{gao2018learning}
Ruohan Gao, Rogerio Feris, and Kristen Grauman.
\newblock Learning to separate object sounds by watching unlabeled video.
\newblock In {\em Proceedings of the European Conference on Computer Vision
  (ECCV)}, pages 35--53, 2018.

\bibitem{gao20192}
Ruohan Gao and Kristen Grauman.
\newblock 2.5 d visual sound.
\newblock In {\em Proceedings of the IEEE/CVF Conference on Computer Vision and
  Pattern Recognition}, pages 324--333, 2019.

\bibitem{garg2021geometry}
Rishabh Garg, Ruohan Gao, and Kristen Grauman.
\newblock Geometry-aware multi-task learning for binaural audio generation from
  video.
\newblock {\em arXiv preprint arXiv:2111.10882}, 2021.

\bibitem{gatys2015neural}
Leon~A Gatys, Alexander~S Ecker, and Matthias Bethge.
\newblock A neural algorithm of artistic style.
\newblock {\em arXiv preprint arXiv:1508.06576}, 2015.

\bibitem{gatys2016image}
Leon~A Gatys, Alexander~S Ecker, and Matthias Bethge.
\newblock Image style transfer using convolutional neural networks.
\newblock In {\em Proceedings of the IEEE conference on computer vision and
  pattern recognition}, pages 2414--2423, 2016.

\bibitem{ghose2020autofoley}
Sanchita Ghose and John~Jeffrey Prevost.
\newblock Autofoley: Artificial synthesis of synchronized sound tracks for
  silent videos with deep learning.
\newblock {\em IEEE Transactions on Multimedia}, 23:1895--1907, 2020.

\bibitem{ghose2021foleygan}
Sanchita Ghose and John~J Prevost.
\newblock Foleygan: Visually guided generative adversarial network-based
  synchronous sound generation in silent videos.
\newblock {\em arXiv preprint arXiv:2107.09262}, 2021.

\bibitem{goodfellow2014generative}
Ian Goodfellow, Jean Pouget-Abadie, Mehdi Mirza, Bing Xu, David Warde-Farley,
  Sherjil Ozair, Aaron Courville, and Yoshua Bengio.
\newblock Generative adversarial nets.
\newblock In {\em Advances in Neural Information Processing Systems}, pages
  2672--2680, 2014.

\bibitem{griffin1984signal}
Daniel Griffin and Jae Lim.
\newblock Signal estimation from modified short-time fourier transform.
\newblock {\em IEEE Transactions on acoustics, speech, and signal processing},
  32(2):236--243, 1984.

\bibitem{hershey2001audio}
John Hershey and Michael Casey.
\newblock Audio-visual sound separation via hidden markov models.
\newblock {\em Advances in Neural Information Processing Systems}, 14, 2001.

\bibitem{hershey2016cnn}
Shawn Hershey, Sourish Chaudhuri, Daniel~PW Ellis, Jort~F Gemmeke, Aren Jansen,
  R~Channing Moore, Manoj Plakal, Devin Platt, Rif~A Saurous, Bryan Seybold,
  et~al.
\newblock Cnn architectures for large-scale audio classification.
\newblock {\em arXiv preprint arXiv:1609.09430}, 2016.

\bibitem{hershey2017cnn}
Shawn Hershey, Sourish Chaudhuri, Daniel P.~W. Ellis, Jort~F. Gemmeke, Aren
  Jansen, Channing Moore, Manoj Plakal, Devin Platt, Rif~A. Saurous, Bryan
  Seybold, Malcolm Slaney, Ron Weiss, and Kevin Wilson.
\newblock Cnn architectures for large-scale audio classification.
\newblock In {\em International Conference on Acoustics, Speech and Signal
  Processing (ICASSP)}. 2017.

\bibitem{hertzmann2001image}
Aaron Hertzmann, Charles~E Jacobs, Nuria Oliver, Brian Curless, and David~H
  Salesin.
\newblock Image analogies.
\newblock In {\em Proceedings of the 28th annual conference on Computer
  graphics and interactive techniques}, pages 327--340, 2001.

\bibitem{huang2018timbretron}
Sicong Huang, Qiyang Li, Cem Anil, Xuchan Bao, Sageev Oore, and Roger~B Grosse.
\newblock Timbretron: A wavenet (cyclegan (cqt (audio))) pipeline for musical
  timbre transfer.
\newblock {\em arXiv preprint arXiv:1811.09620}, 2018.

\bibitem{iashin2021taming}
Vladimir Iashin and Esa Rahtu.
\newblock Taming visually guided sound generation.
\newblock {\em arXiv preprint arXiv:2110.08791}, 2021.

\bibitem{iashin2022sparse}
Vladimir Iashin, Weidi Xie, Esa Rahtu, and Andrew Zisserman.
\newblock Sparse in space and time: Audio-visual synchronisation with trainable
  selectors.
\newblock {\em arXiv preprint arXiv:2210.07055}, 2022.

\bibitem{isola2017image}
Phillip Isola, Jun-Yan Zhu, Tinghui Zhou, and Alexei~A Efros.
\newblock Image-to-image translation with conditional adversarial networks.
\newblock In {\em Proceedings of the IEEE conference on computer vision and
  pattern recognition}, pages 1125--1134, 2017.

\bibitem{johnson2016perceptual}
Justin Johnson, Alexandre Alahi, and Li Fei-Fei.
\newblock Perceptual losses for real-time style transfer and super-resolution,
  2016.

\bibitem{kingma2015adam}
Diederik Kingma and Jimmy Ba.
\newblock Adam: A method for stochastic optimization.
\newblock International Conference on Learning Representation, 2015.

\bibitem{kingma2013auto}
Diederik~P Kingma and Max Welling.
\newblock Auto-encoding variational bayes.
\newblock {\em arXiv preprint arXiv:1312.6114}, 2013.

\bibitem{koepke2020sight}
A~Sophia Koepke, Olivia Wiles, Yael Moses, and Andrew Zisserman.
\newblock Sight to sound: An end-to-end approach for visual piano
  transcription.
\newblock In {\em ICASSP 2020-2020 IEEE International Conference on Acoustics,
  Speech and Signal Processing (ICASSP)}, pages 1838--1842. IEEE, 2020.

\bibitem{kumar2019melgan}
Kundan Kumar, Rithesh Kumar, Thibault de Boissiere, Lucas Gestin, Wei~Zhen
  Teoh, Jose Sotelo, Alexandre de Br{\'e}bisson, Yoshua Bengio, and Aaron~C
  Courville.
\newblock Melgan: Generative adversarial networks for conditional waveform
  synthesis.
\newblock {\em Advances in neural information processing systems}, 32, 2019.

\bibitem{langlois2014inverse}
Timothy~R Langlois and Doug~L James.
\newblock Inverse-foley animation: synchronizing rigid-body motions to sound.
\newblock {\em ACM Transactions on Graphics (TOG)}, 33(4):41, 2014.

\bibitem{lee2021sound}
Seung~Hyun Lee, Wonseok Roh, Wonmin Byeon, Sang~Ho Yoon, Chan~Young Kim, Jinkyu
  Kim, and Sangpil Kim.
\newblock Sound-guided semantic image manipulation.
\newblock {\em arXiv preprint arXiv:2112.00007}, 2021.

\bibitem{li2022learning}
Tingle Li, Yichen Liu, Andrew Owens, and Hang Zhao.
\newblock Learning visual styles from audio-visual associations.
\newblock {\em arXiv}, 2022.

\bibitem{nistal2020drumgan}
Javier Nistal, Stefan Lattner, and Gael Richard.
\newblock Drumgan: Synthesis of drum sounds with timbral feature conditioning
  using generative adversarial networks.
\newblock {\em arXiv preprint arXiv:2008.12073}, 2020.

\bibitem{owens2018audio}
Andrew Owens and Alexei~A Efros.
\newblock Audio-visual scene analysis with self-supervised multisensory
  features.
\newblock {\em European Conference on Computer Vision (ECCV)}, 2018.

\bibitem{owens2015visually}
Andrew Owens, Phillip Isola, Josh McDermott, Antonio Torralba, Edward~H
  Adelson, and William~T Freeman.
\newblock Visually indicated sounds.
\newblock {\em CVPR}, 2016.

\bibitem{owens2016visually}
Andrew Owens, Phillip Isola, Josh McDermott, Antonio Torralba, Edward~H
  Adelson, and William~T Freeman.
\newblock Visually indicated sounds.
\newblock In {\em Computer Vision and Pattern Recognition (CVPR)}, 2016.

\bibitem{prajwal2020learning}
KR Prajwal, Rudrabha Mukhopadhyay, Vinay~P Namboodiri, and CV Jawahar.
\newblock Learning individual speaking styles for accurate lip to speech
  synthesis.
\newblock In {\em Proceedings of the IEEE/CVF Conference on Computer Vision and
  Pattern Recognition}, pages 13796--13805, 2020.

\bibitem{radford2019language}
Alec Radford, Jeffrey Wu, Rewon Child, David Luan, Dario Amodei, Ilya
  Sutskever, et~al.
\newblock Language models are unsupervised multitask learners.
\newblock 2019.

\bibitem{raffel2014mir_eval}
Colin Raffel, Brian McFee, Eric~J Humphrey, Justin Salamon, Oriol Nieto, Dawen
  Liang, Daniel~PW Ellis, and C~Colin Raffel.
\newblock Mir\_eval: A transparent implementation of common mir metrics.
\newblock In {\em ISMIR}, pages 367--372, 2014.

\bibitem{ramesh2021zero}
Aditya Ramesh, Mikhail Pavlov, Gabriel Goh, Scott Gray, Chelsea Voss, Alec
  Radford, Mark Chen, and Ilya Sutskever.
\newblock Zero-shot text-to-image generation.
\newblock In {\em International Conference on Machine Learning}, pages
  8821--8831. PMLR, 2021.

\bibitem{sainburg2020finding}
Tim Sainburg, Marvin Thielk, and Timothy~Q Gentner.
\newblock Finding, visualizing, and quantifying latent structure across diverse
  animal vocal repertoires.
\newblock {\em PLoS computational biology}, 16(10):e1008228, 2020.

\bibitem{shechtman2007matching}
Eli Shechtman and Michal Irani.
\newblock Matching local self-similarities across images and videos.
\newblock In {\em 2007 IEEE Conference on Computer Vision and Pattern
  Recognition}, pages 1--8. IEEE, 2007.

\bibitem{su2020multi}
Kun Su, Xiulong Liu, and Eli Shlizerman.
\newblock Multi-instrumentalist net: Unsupervised generation of music from body
  movements.
\newblock {\em arXiv preprint arXiv:2012.03478}, 2020.

\bibitem{su2021does}
Kun Su, Xiulong Liu, and Eli Shlizerman.
\newblock How does it sound?
\newblock {\em Advances in Neural Information Processing Systems}, 34, 2021.

\bibitem{suris2022time}
Dídac Surís, Carl Vondrick, Bryan Russell, and Justin Salamon.
\newblock It's time for artistic correspondence in music and video.
\newblock {\em CVPR}, 2022.

\bibitem{tran2018}
Du Tran, Heng Wang, Lorenzo Torresani, Jamie Ray, Yann LeCun, and Manohar
  Paluri.
\newblock A closer look at spatiotemporal convolutions for action recognition.
\newblock In {\em 2018 IEEE/CVF Conference on Computer Vision and Pattern
  Recognition}, pages 6450--6459, 2018.

\bibitem{tzinis2020into}
Efthymios Tzinis, Scott Wisdom, Aren Jansen, Shawn Hershey, Tal Remez,
  Daniel~PW Ellis, and John~R Hershey.
\newblock Into the wild with audioscope: Unsupervised audio-visual separation
  of on-screen sounds.
\newblock {\em arXiv preprint arXiv:2011.01143}, 2020.

\bibitem{ulyanov2016audio}
Dmitry Ulyanov.
\newblock Audio texture synthesis and style transfer.
\newblock
  \url{https://dmitryulyanov.github.io/audio-texture-synthesis-and-style-transfer},
  2016.

\bibitem{van2001foleyautomatic}
Kees Van Den~Doel, Paul~G Kry, and Dinesh~K Pai.
\newblock Foleyautomatic: physically-based sound effects for interactive
  simulation and animation.
\newblock In {\em Proceedings of the 28th annual conference on Computer
  graphics and interactive techniques}, pages 537--544. ACM, 2001.

\bibitem{van2017neural}
Aaron Van Den~Oord, Oriol Vinyals, et~al.
\newblock Neural discrete representation learning.
\newblock {\em Advances in neural information processing systems}, 30, 2017.

\bibitem{vaswani2017attention}
Ashish Vaswani, Noam Shazeer, Niki Parmar, Jakob Uszkoreit, Llion Jones,
  Aidan~N Gomez, {\L}ukasz Kaiser, and Illia Polosukhin.
\newblock Attention is all you need.
\newblock In {\em Advances in neural information processing systems}, pages
  5998--6008, 2017.

\bibitem{vaswani2017}
Ashish Vaswani, Noam Shazeer, Niki Parmar, Jakob Uszkoreit, Llion Jones,
  Aidan~N Gomez, Lukasz Kaiser, and Illia Polosukhin.
\newblock Attention is all you need.
\newblock In {\em Neural Information Processing Systems (NIPS)}, 2017.

\bibitem{verma2018neural}
Prateek Verma and Julius~O Smith.
\newblock Neural style transfer for audio spectograms.
\newblock {\em arXiv preprint arXiv:1801.01589}, 2018.

\bibitem{wang2021calls}
Yu Wang, Nicholas~J Bryan, Justin Salamon, Mark Cartwright, and Juan~Pablo
  Bello.
\newblock Who calls the shots? rethinking few-shot learning for audio.
\newblock In {\em 2021 IEEE Workshop on Applications of Signal Processing to
  Audio and Acoustics (WASPAA)}, pages 36--40. IEEE, 2021.

\bibitem{wang2020few}
Yu Wang, Justin Salamon, Nicholas~J Bryan, and Juan~Pablo Bello.
\newblock Few-shot sound event detection.
\newblock In {\em ICASSP 2020-2020 IEEE International Conference on Acoustics,
  Speech and Signal Processing (ICASSP)}, pages 81--85. IEEE, 2020.

\bibitem{xu2021visually}
Xudong Xu, Hang Zhou, Ziwei Liu, Bo Dai, Xiaogang Wang, and Dahua Lin.
\newblock Visually informed binaural audio generation without binaural audios.
\newblock In {\em Proceedings of the IEEE/CVF Conference on Computer Vision and
  Pattern Recognition}, pages 15485--15494, 2021.

\bibitem{yang2020telling}
Karren Yang, Bryan Russell, and Justin Salamon.
\newblock Telling left from right: Learning spatial correspondence of sight and
  sound.
\newblock In {\em Proceedings of the IEEE/CVF Conference on Computer Vision and
  Pattern Recognition}, pages 9932--9941, 2020.

\bibitem{zhang2017real}
Richard Zhang, Jun-Yan Zhu, Phillip Isola, Xinyang Geng, Angela~S Lin, Tianhe
  Yu, and Alexei~A Efros.
\newblock Real-time user-guided image colorization with learned deep priors.
\newblock {\em arXiv preprint arXiv:1705.02999}, 2017.

\bibitem{zhang2021repetitive}
Yunhua Zhang, Ling Shao, and Cees~GM Snoek.
\newblock Repetitive activity counting by sight and sound.
\newblock In {\em Proceedings of the IEEE/CVF Conference on Computer Vision and
  Pattern Recognition}, pages 14070--14079, 2021.

\bibitem{zhao2018sound}
Hang Zhao, Chuang Gan, Andrew Rouditchenko, Carl Vondrick, Josh McDermott, and
  Antonio Torralba.
\newblock The sound of pixels.
\newblock In {\em Proceedings of the European conference on computer vision
  (ECCV)}, pages 570--586, 2018.

\bibitem{zhou2018visual}
Yipin Zhou, Zhaowen Wang, Chen Fang, Trung Bui, and Tamara~L Berg.
\newblock Visual to sound: Generating natural sound for videos in the wild.
\newblock In {\em Proceedings of the IEEE conference on computer vision and
  pattern recognition}, pages 3550--3558, 2018.

\bibitem{zhu2017unpaired}
Jun-Yan Zhu, Taesung Park, Phillip Isola, and Alexei~A Efros.
\newblock Unpaired image-to-image translation using cycle-consistent
  adversarial networks.
\newblock 2017.

\end{thebibliography}
}
\clearpage
\appendix
\supparxiv{
\setcounter{page}{1}
\twocolumn[{%
\renewcommand\twocolumn[1][]{#1}%
\begin{center}
    \vspace{-2.0em}
    {\bf \large Supplementary Material:\\Conditional Generation of Audio from Video via Foley Analogies}
    \vspace{2.0em}
\end{center}

}]

}{}

\renewcommand{\thesection}{A.\arabic{section}}
\setcounter{section}{0}

\section{VQGAN codebook loss}
\label{appendix:loss}
We follow the formulation of  Iashin and Rahtu~\cite{iashin2021taming}, which adapts the VQGAN loss~\cite{esser2021taming,van2017neural} to spectrograms. The loss $\mathcal{L}$ is composed of three individual components: the reconstruction and codebook loss~\cite{kingma2013auto,van2017neural} $\mathcal{L}_{\mathit{VQVAE}}$, the perceptual loss~\cite{johnson2016perceptual} $\mathcal{L}_{perceptual}$, and a patch-based discriminator loss~\cite{isola2017image} $\mathcal{L}_{disc}$. We present the final loss we used to learn the codebook:
\begin{equation}
    \mathcal{L}_{\mathit{VQGAN}} = \mathcal{L}_{\mathit{VQVAE}} + \mathcal{L}_{perceptual} + \mathcal{L}_{disc}
    \label{eq:vqgan}
\end{equation} We describe these losses next. 

\mypar{Reconstruction and codebook loss.} Given the input waveform $\ba$ and its mel spectrogram $\mathbf{S} = \mathrm{MSTFT}(\ba) \in \mathbb{R}^{T \times F}$, where $T$ and $F$ are the dimension for time and frequency, the encoder $E$ will encode $\mathbf{S}$ into embeddings $\hat{\bz}_{tf}$ and corresponding quantized code $q(\hat{\bz}_{tf}) = {\bz}_{tf}$. The decoder $D$ will then reconstruct ${\bz}_{tf}$ to a mel spectrogram $\hat {\mathbf S} = D(q(E(\mathrm{MSTFT}(\ba))))$. Following VQVAE, we minimize the reconstruction loss between $\mathbf{S}$ and $\hat{\mathbf S}$ as well as the {\em codebook loss} between $\hat{\bz}_{tf}$ and ${\bz}_{tf}$: 
\begin{equation}
    \mathcal{L}_{\mathit{VQVAE}} = \underbrace{||\mathbf{S} - \hat{\mathbf S}||}_{\text{recon. loss}} + \underbrace{|| \hat{\bz}_{tf} - \text{sg}[{\bz}_{tf}] ||^2_2 + || \text{sg}[\hat{\bz}_{tf}] - {\bz}_{tf} ||^2_2}_{\text{codebook loss}}.
    \label{eq:vqvae}
\end{equation}
where $\text{sg}$ is the stop gradient operation.

\mypar{Perceptual loss.} We use the off-the-shelf VGGish-ish model~\cite{iashin2021taming}, a variation of the VGGish model with VGG-16 backbone trained on the VGGSound dataset, to evaluate the perceptual loss. For the $i^{\text{th}}$ level of features of the original and the reconstructed mel spectrogram, $\mathbf{S}^i$ and $\hat{\mathbf{S}}^i$ respectively, the corresponding perceptual loss is given by their squared L2 distance:
\begin{equation}
    \mathcal{L}_{perceptual} = \sum_{i=1}^{N} \frac{1}{F^i T^i}|| \mathbf{S}^i - \hat{\mathbf{S}}^i ||^2_2
    \label{eq:perceptual},
\end{equation}
where $N$ is the number of layers we selected to calculate the perceptual loss. We use $N = 5$ layers in our model, selected as in~\cite{iashin2021taming}.

\mypar{Patch-based discriminator loss.} We use the discriminator loss introduced by Isola \etal~\cite{isola2017image}:
\begin{equation}
    \mathcal{L}_{disc} = \log \mathcal{D}(\mathbf{S}) + \log(1 - \mathcal{D}(\hat{\mathbf{S}})),
    \label{eq:adv}
\end{equation}
where $\mathcal{D}$ is applied to a fully convolutional network at multiple scales as a discriminator.

\section{Dataset}
\label{appendix:dataset}
All data comes from the {\em Greatest Hits} dataset, a lab-collected dataset from Owens \etal~\cite{owens2015visually}, and the {\em CountixAV}~\cite{zhang2021repetitive} dataset. The {\em Greatest Hits} dataset is composed of 977 videos~(11 hours) of a drumstick interacting with different objects in the scenes. The {\em CountixAV}~\cite{zhang2021repetitive} dataset is composed of 1483 videos~(4.1 hours) of repeated actions including bouncing ball, skipping rope, and other actions in realistic scenarios.

To evaluate our conditional Foley generation task, we randomly sample 2-sec video clips from the test videos in the {\em Greatest Hits}~\cite{owens2015visually} dataset as input videos. We randomly pair each silent input video with 3 conditional audio-visual clips from different test videos. We obtain the conditional Foley evaluation set of 582 input-condition pairs with 17 different materials and two action types.

We also evaluate our model on the {\em CountixAV}~\cite{zhang2021repetitive} dataset by randomly choosing 2 seconds clips from the videos. We apply a noise reduction algorithm~\cite{sainburg2020finding} to improve the sound quality before training and evaluation. Considering the lack of the permission of the videos in the {\em CountixAV}~\cite{zhang2021repetitive} dataset, we demonstrate the result qualitatively on the publicly-sourced videos with proper permissions. We provide the credit to those videos in the \cref{appendix:credit}

\section{Implementation details} 
\label{appendix:implementation}
To train our complete model, we first train the VQGAN, then train the audio predictor using its learned code. We trained the VQGAN for approximately 400 epochs with a batch size of 32 and a learning rate of $1.44 \times 10^{-4}$ using Adam~\cite{kingma2015adam}. We trained the transformer for 50 epochs with a batch size of 8 and a learning rate of $1.6 \times 10^{-4}$ using 4 NVIDIA A40 GPUs. The training of the VQGAN takes approximately 5 days and the training of the transformer takes roughly 20 hours.

Our model operates on 2 sec.\ video clips for both the conditional and input videos, at a 15Hz video sampling rate and 22.05 KHz audio sampling rate. The VQGAN codebook encoder produces a $12 \times 5$ time-frequency grid from a mel spectrogram, which is converted from a waveform using 80 mel bins and 1,024 Fourier filters. We use $d = 256$ for the codebook embedding dimension. The 2-sec.\ videos~(30 frames) are randomly cropped and resized to $112 \times 112$, and are represented as 30 1024-dimensional feature vectors, obtained by performing a $1 \times 1$ convolution on the ResNet feature map. During training, we apply data augmentation in the form of frequency and temporal masking to the spectrogram prior to extracting the clips. The model is trained on both the {\em Greatest Hits}~\cite{owens2015visually} dataset and the {\em CountixAV}~\cite{zhang2021repetitive} dataset in the same manner.

During inference, we set the re-ranking tolerance to be $\tau=0.2$.

\begin{figure*}[ht!]
    \centering
    \vspace{-1mm}
    \includegraphics[width=0.95\linewidth]{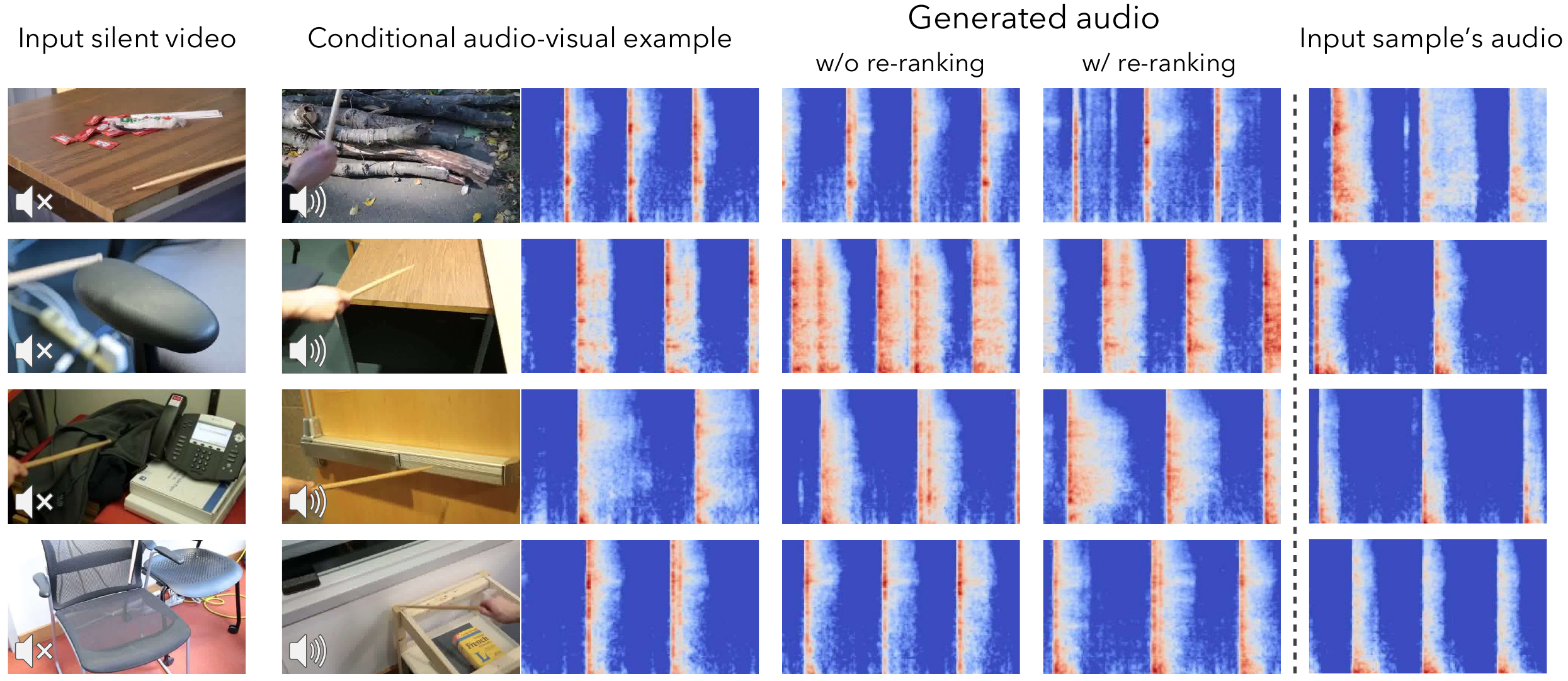}
    \caption{{\bf Re-reranked comparison results}. We present 4 pairs of results to compare the effect of re-ranking qualitatively. }  \vspace{-1mm}
    \label{fig:rerank}
\end{figure*}   

\section{Onset transfer baseline training details}
\label{appendix:onset}
For predicting onsets given a video, we used the same variation of ResNet (2+1)D-18~\cite{tran2018} visual network as our main model (i.e., after removing temporal striding). Our model outputs a vector of predictions using a fully connected layer after pooling (one onset prediction per input frame). We obtain the ground truth onset according to the timing label provided in the {\em Greatest Hits}~\cite{owens2015visually} dataset by aligning it to the closest frame. We use a binary cross entropy loss, penalizing onset predictions that occur at incorrect times. Since each input video can have more than one sound event, we set the weight of each onset in the loss from the video equally according to the total number of onsets in the input so that the weight sums to one. The configuration of video and audio is the same as our model, we use a frame rate of 15Hz and an audio sampling rate of 22.05kHz. We train the model for 100 epochs with a batch size of $12$ clips of 2 seconds and a learning rate of $1\times 10^{-4}$ using Adam~\cite{kingma2015adam}. The model is trained on a single NVIDIA RTX-2080 GPU.

The model is used to detect the onset in both input and conditional videos. We then randomly copy-and-paste the sound from the conditional video at onset timings. We included the onset transfer method to help analyze the generative models on Greatest Hits~[\textcolor{green}{40}]. It simply detects onsets and copy-pastes sounds from the conditional example. By design, it (trivially) obtains near-perfect performance on the {\em material} metric when there is only one material and also performs well on onset metrics  {\em because it is directly trained on onset labels}. We emphasize that this baseline is not generative and, in fact, we show that this method will fail when the action is different in the input and condition pairs (column 6 in \cref{tab:VGGishTestMerged}). Qualitatively, we have found that it often completely fails on CountixAV~[\textcolor{green}{60}] since there are no clear onsets. The number of onsets is rarely correct and the method fails due to background noise (\eg row 1 in \cref{fig:qualitatve}).

\section{Sound classifier training details for quantitative experiment}
\label{appendix:soundclassifier}
We finetune two pretrained VGGish classifiers~\cite{hershey2016cnn, hershey2017cnn} to predict the action or the material presented in the video based on the label provided in the {\em Greatest Hits}~\cite{owens2015visually} dataset. With the same input video settings, we train both of the models with an early stopping criteria using Adam~\cite{kingma2015adam}. The model is trained on a single NVIDIA RTX-2080 GPU with a learning rate of $10^{-4}$ and batch size $32$ video clips of 2 seconds. The trained model obtains a validation accuracy of $75.6\%$ on the material task and $92.0\%$ accuracy on the action prediction task.

\begin{table}[t]
\centering
\resizebox{\linewidth}{!}
{
\begin{tabular}{lccc}
\toprule
\multirow{4}{*}{Model} & \multicolumn{3}{c}{Window size}                 \\ \cmidrule{2-4} 
                       & 0.1-sec. & 0.2-sec. & Avg. \\
                       & AP (\%)  & AP(\%)   & AP(\%)                 \\ \midrule
Style transfer\textsuperscript{$\ast$}~\cite{ulyanov2016audio,gatys2015neural}        & 46.9             & 46.7  & 58.3             \\
Onset transfer         & \textbf{71.9}             & \textbf{78.4}  & \textbf{76.5}                \\ \midrule
SpecVQGAN\cite{iashin2021taming}              & 59.3             & 65.5  & 64.1                \\
Ours - No cond.        & 59.3             & 73.2  & 69.8                \\
Ours - Base            & \textbf{60.0}             & \textbf{74.0}  & \textbf{70.0}                \\ \bottomrule
\end{tabular}
}
\caption{{\bf Onset synchronization window size evaluation.} We measure the average precision of onset predictions that are within different windows size of 0.1, and 0.2 seconds. We also measure the averaged AP score under different window sizes (from 0.10s to 0.25s with a step of 0.05s).}
\label{tab:windowsize}

\end{table}
\section{Onset detection experiment with different window size}

We chose 0.1 seconds as the size of the detection window for the onset detection experiment following the standard value used for audio onset detection per \texttt{mir\_eval}~\cite{raffel2014mir_eval} and the {\em Greatest Hits}~\cite{owens2015visually}. Alternatively, we have also experimented with different methods with alternative window sizes. We found that (\cref{tab:windowsize}) our method outperforms all the other generative baselines in the additional experiment. Meanwhile, the gap between our method and the onset transfer baseline is reduced when the window size is enlarged or averaged over different window sizes.

\begin{figure}[t]
    \centering
    \includegraphics[width=1.0\columnwidth]{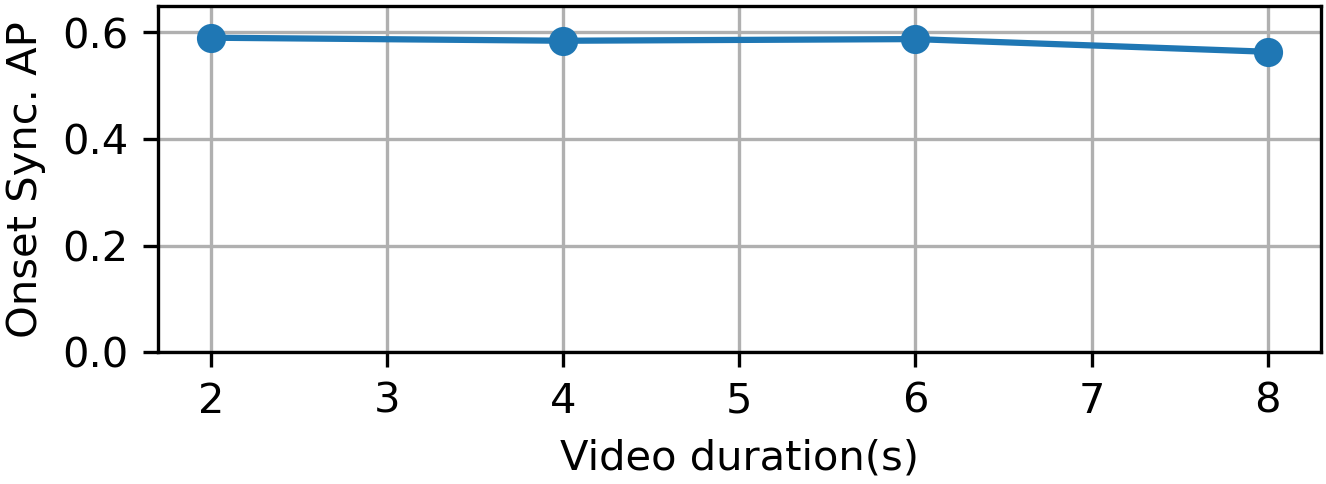}
    \vspace{-4mm}
    \caption{{\bf Automated onset sync. evaluation for different durations}. We evaluate the performance of the onset synchronization for videos of different durations.}
    \label{fig:longer_onset_ap}
    \vspace{-3mm}
\end{figure}
\section{Generation of longer audio}

Following Iashin \etal~\cite{iashin2021taming}, we generate longer audio clips using a 2-sec. sliding window. During the process, we always keep the token to be generated in the center of all the visible tokens before feeding the tokens into the transformer. We have evaluated the onset sync.\ performance for the {\em Greatest Hits} dataset~[\textcolor{green}{40}] for 454 pairs of videos. From \cref{fig:longer_onset_ap}, we note that our model continues to obtain strong performance for videos up to 6 sec. We provide examples of the generation of 4-sec., 6-sec., and 8-sec. in our \href{https://xypb.github.io/CondFoleyGen/}{project webpage}.

\section{Re-ranking qualitative result}

We notice that the re-ranking improves the synchronization performance through the qualitative result in \cref{fig:rerank}. Note the generated sound with re-ranking shows a much better synchronization performance in Row 2 and 3 in \cref{fig:rerank}.

\section{Human study details}

For the human study, we recruited 609 participants from Amazon Mechanical Turk, in total. We filtered out 233 pieces of feedback according to the answers the participants provided to a sentinel example, where one of the videos is attached with a clearly wrong sound that neither match with the action in the video, nor provide a timbre close to the chosen conditional video. We obtained 376 effective feedback in the end.

We provided them with the following instructions:
\begin{itemize}[leftmargin=*,topsep=1pt, noitemsep]
    \item Please use headphones for the test. You may hear some harsh or dissonant sounds, so please make sure to adjust your device's volume before the test.
    \item The task should take approximately 15 minutes to complete.
    \item You will take part in an experiment involving visual and hearing perception. To complete this task, you will need to watch and listen to 21 groups of videos and answer two questions for each group. Each group consists of the following videos (all videos have audio):
    \begin{itemize}
        \item One input video: Video \#1
        \item Two output videos: Video \#2 and Video \#3
    \end{itemize}
    \item Your task is to answer the two questions at the bottom:
    \begin{itemize}
        \item One input video: Video \#1. In which output Video (\#2 or \#3) is the audio most synchronized with the action in the video?
        \item In which output Video (\#2 or \#3) does the audio sound most like the object or material in Video \#1 according to the action in the output Video?
    \end{itemize}
    \item You will complete a short practice of 5 groups of videos (about 3 minutes long) before starting the main task, so that you can get familiar with the interface.
\end{itemize}

\begin{figure*}[ht!]
    \centering
    \includegraphics[width=0.95\linewidth]{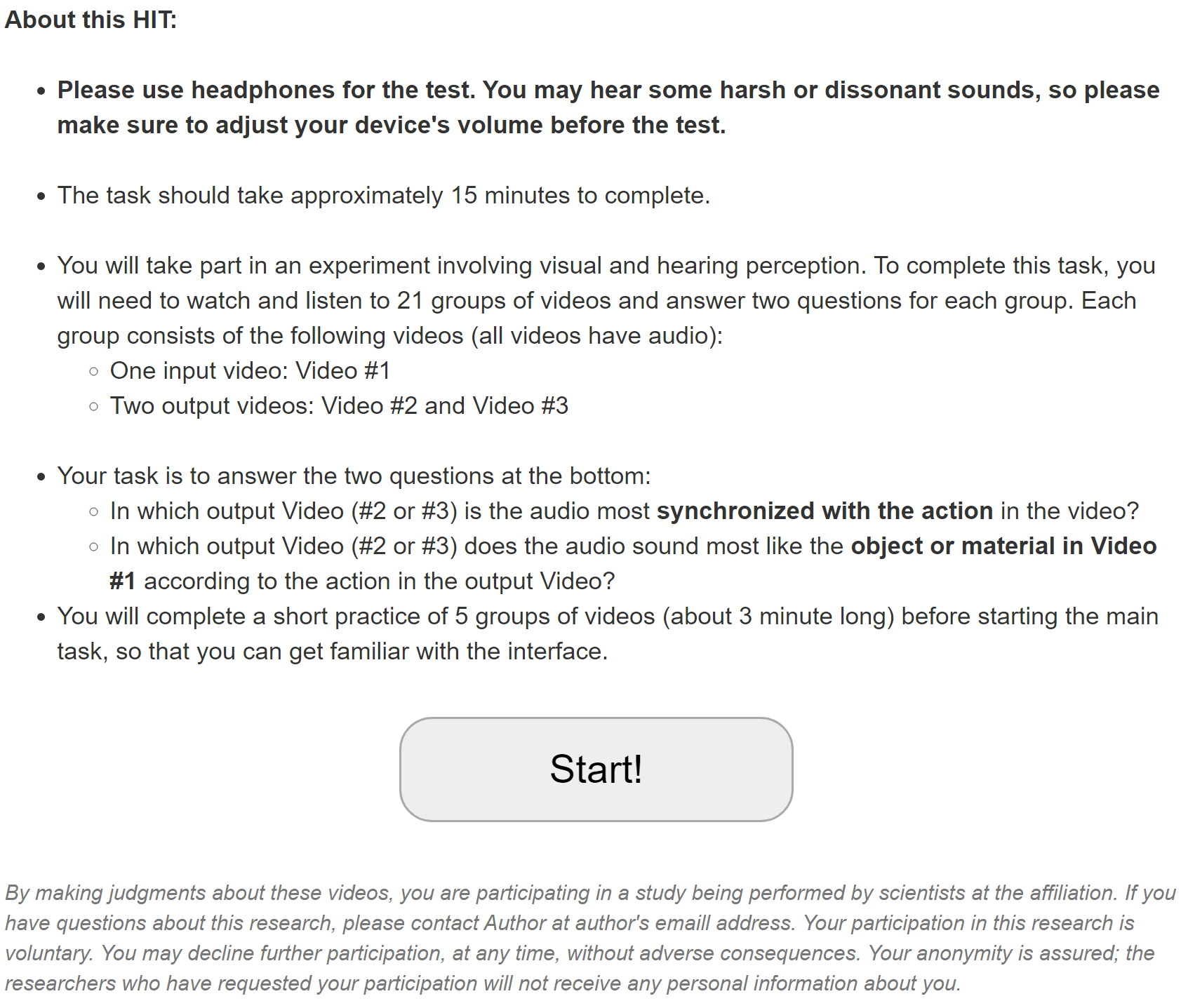}
    \caption{{\bf Instruction page of the AMT test}. We present a screenshot of the instruction page that the participants will see at the beginning of the test. The sensitive information is removed from the image. }  \vspace{-1mm}
    \label{fig:instruction}
\end{figure*}

\begin{figure*}[ht!]
    \centering
    \vspace{-1mm}
    \includegraphics[width=0.95\linewidth]{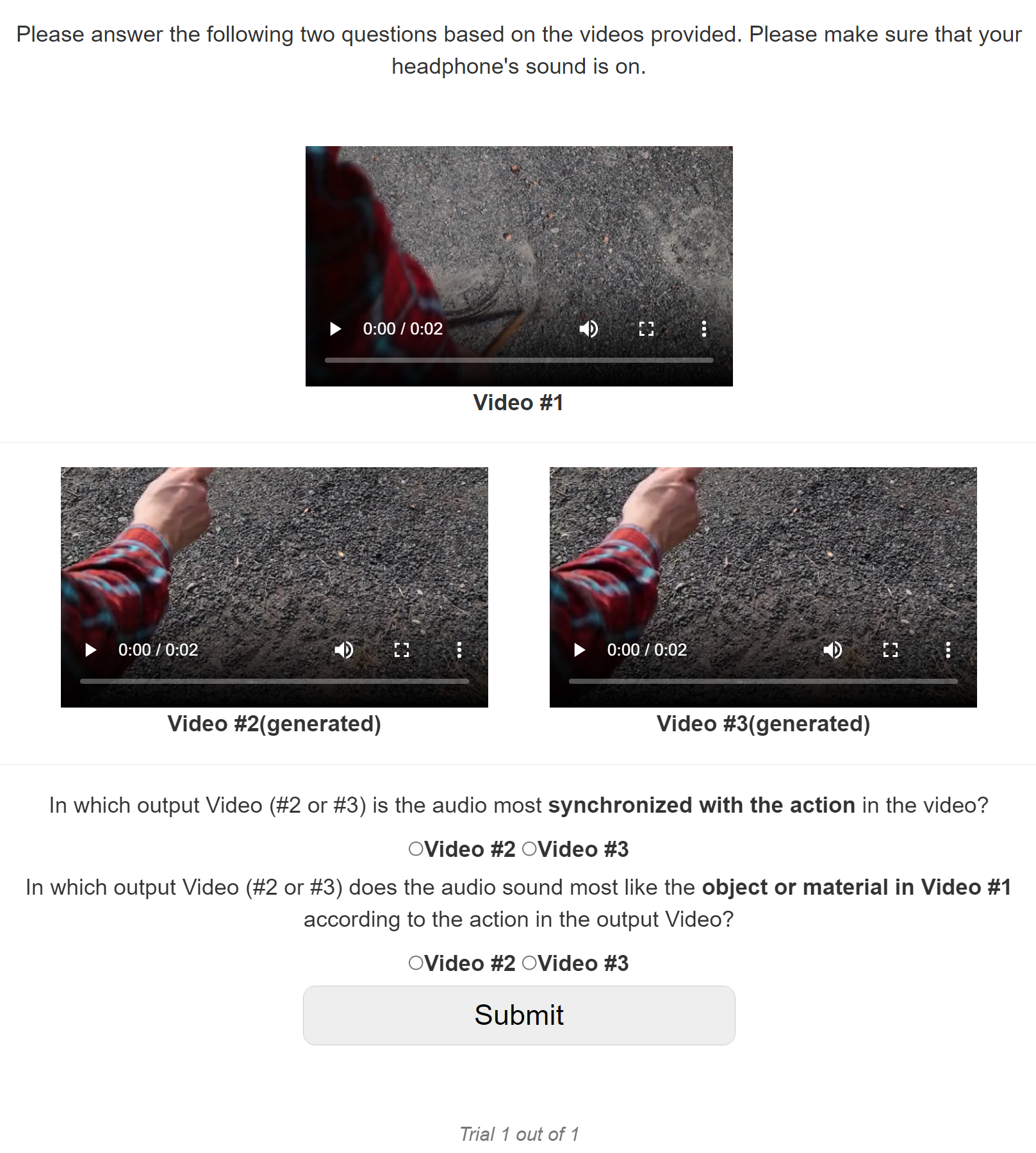}
    \caption{{\bf Test page of the AMT test}. We present a screenshot of the main test page that the participants will see during the test. The participant need to answer both of the questions before moving to the next set of videos. Clicking on the ``submit'' button will navigate the participant to the next question. }  \vspace{-1mm}
    \label{fig:test_body}
\end{figure*}

We also provide the screenshot of the instruction page and main test page that the participant will see during the test in \cref{fig:instruction} and \cref{fig:test_body}.


\section{Randomly selected results}
\begin{figure*}[ht!]
    \centering
    \vspace{-1mm}
    \includegraphics[width=0.95\linewidth]{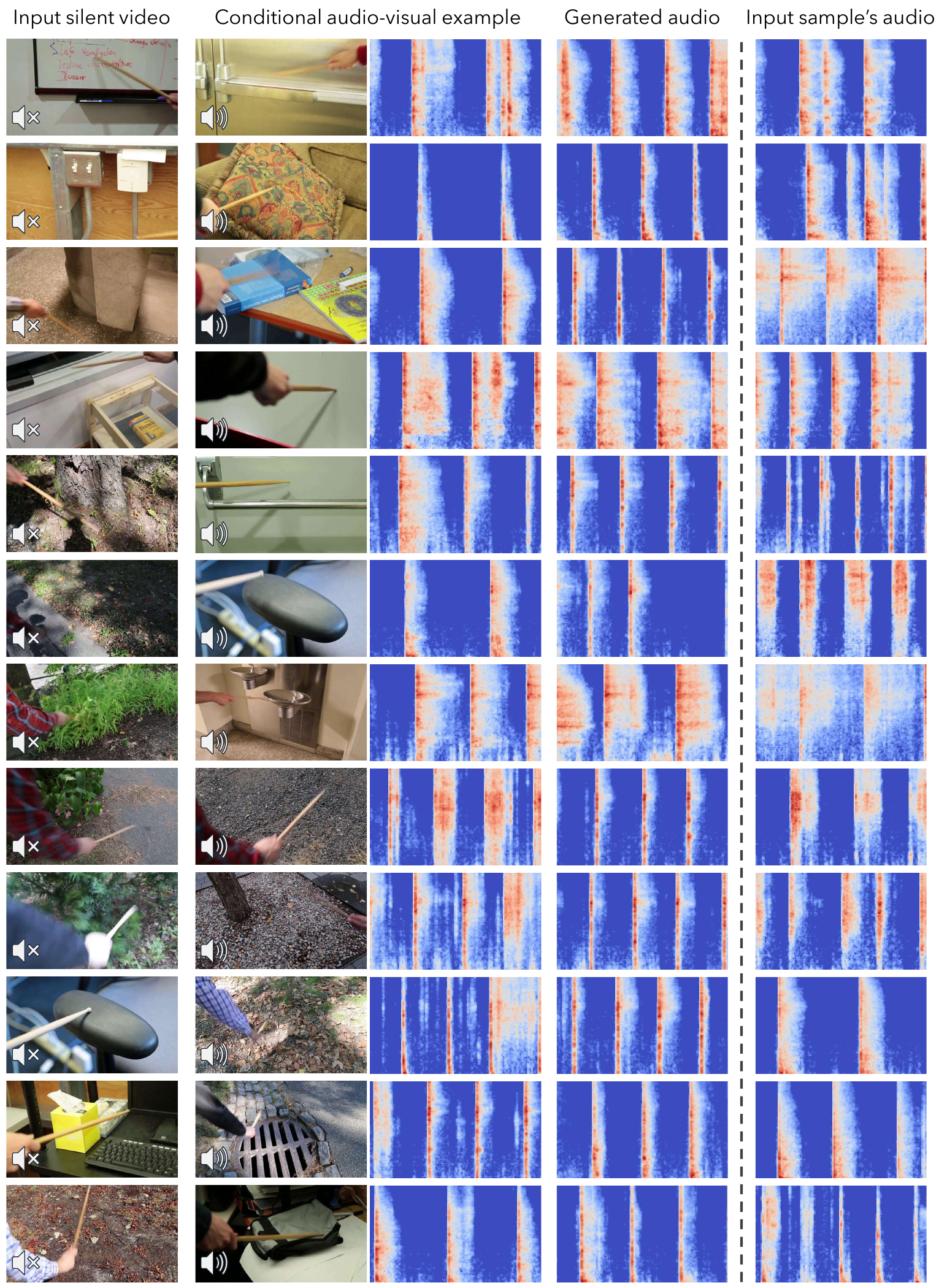}
    \caption{{\bf Randomly selected results}. We present 12 results generated by our model. Please refer to the video to hear the results. }  \vspace{-1mm}
    \label{fig:random}
\end{figure*}

We provide randomly selected results in \cref{fig:random} (this supplemental). Please also see the provided video. It is notable that the generated audio provides a timbre very close to the conditional video in most situations. Meanwhile, the model can also generate a sound match with the timing of the actions in the input video usually.

\section{Video Credits}
\label{appendix:credit}
We have obtained permission to use and edit the videos from the {\em Greatest Hits} dataset, and we have also recorded some of the videos ourselves. We provide here the credit for the publicly sourced and licensed videos that appear in the paper and the supplement.
\small
\begin{enumerate}
    \item fotorealis - \url{https://stock.adobe.com/video/bambino-suona-bongo-percussioni-per-musicoterapia/536371955} - Adobe Stock Extended license.
    \item FreeStockFootageClub - \url{https://www.youtube.com/watch?v=zI1_fAtpHQc} - YouTube Creative Commons CC BY license.
    \item Ignat Gorazd - \url{https://www.youtube.com/watch?v=1_ckbCU5aQs} - YouTube Creative Commons CC BY license.
    \item kriista - \url{https://www.youtube.com/watch?v=6d1YS7fdBK4} - YouTube Creative Commons CC BY license.
    \item Over \& Out - \url{https://www.youtube.com/watch?v=SExIpBIBj_k} - YouTube Creative Commons CC BY license.
    \item Over \& Out - \url{https://www.youtube.com/watch?v=XxmZxM8AtUc} - YouTube Creative Commons CC BY license.
    \item Percussion Play - \url{https://www.youtube.com/watch?v=fcjfKvdkJyI} - YouTube Creative Commons CC BY license.
    \item Percussion Play - \url{https://www.youtube.com/watch?v=xcUyiXt0gjo} - YouTube Creative Commons CC BY license.
    \item PhotoSerg - \url{https://stock.adobe.com/video/kid-juggles-the-ping-pong-ball-green-screen/99378579} - Adobe Stock Extended license.
    \item PUSAT E-PEMBELAJARAN UMS - \url{https://www.youtube.com/watch?v=S6TkbV4B4QI} - YouTube Creative Commons CC BY license.
    \item Suliman Razvan - \url{https://stock.adobe.com/video/christian-monk-hitting-a-large-wooden-piece-toaca-with-little-wooden-hammer-to-summon-the-breathren-to-prayer/93378558} - Adobe Stock Extended license.
    \item Thomas Cremier - \url{https://www.youtube.com/watch?v=GFmuVBiwz6k} - YouTube Creative Commons CC BY license.
\end{enumerate}

\end{document}